\documentclass[letterpaper]{article} 
\usepackage[preprint]{aaai2027}  
\usepackage[hyphens]{url}  
\usepackage{graphicx} 
\usepackage{natbib}  
\usepackage{caption} 
\usepackage{lipsum} 
\usepackage{algorithm}
\usepackage{algorithmic}

\usepackage{newfloat}
\usepackage{listings}
\DeclareCaptionStyle{ruled}{labelfont=normalfont,labelsep=colon,strut=off} 
\floatstyle{ruled}
\newfloat{listing}{tb}{lst}{}
\floatname{listing}{Listing}

\usepackage{booktabs}
\usepackage{colortbl}
\usepackage{multirow}
\newcolumntype{L}[1]{>{\raggedright\arraybackslash}p{#1}}
\newcolumntype{C}[1]{>{\centering\arraybackslash}p{#1}}

\definecolor{bestgreen}{RGB}{220, 240, 210}
\definecolor{secondyellow}{RGB}{255, 249, 210}
\definecolor{citeblue}{RGB}{40, 100, 180}

\usepackage{soul}

\title{Prox: Training-Free FFN Activation Sparsity via Approximate Intermediate-Channel Salience in LLMs}
\author{
  Jinyi Liu\textsuperscript{\rm 1,\rm 2},
  Wei Chen\textsuperscript{\rm 1,\rm 2}\corresponding,
  Pengyu Chen\textsuperscript{\rm 1,\rm 2},
  Xinyi Yuan\textsuperscript{\rm 1,\rm 2},
  Minghe Bai\textsuperscript{\rm 3},
  Guoquan Wu\textsuperscript{\rm 1,\rm 2},
  Jun Wei\textsuperscript{\rm 1,\rm 2}
}
\affiliations{
  \textsuperscript{\rm 1}Institute of Software, Chinese Academy of Sciences\\
  \textsuperscript{\rm 2}University of Chinese Academy of Sciences\\
  \textsuperscript{\rm 3}School of Computer Science, Nanjing University of Posts and Telecommunications\\

  liujinyi24@otcaix.iscas.ac.cn,
  wchen@otcaix.iscas.ac.cn,\\
  chenpengyu23@otcaix.iscas.ac.cn,
  yuanxinyi22@otcaix.iscas.ac.cn,\\
  baiminghe24@otcaix.iscas.ac.cn,
  gqwu@otcaix.iscas.ac.cn,
  wj@otcaix.iscas.ac.cn
}

\begin{document}

\maketitle

\begin{abstract}
  Feed-forward networks (FFNs) dominate memory traffic and computation in large language model (LLM) inference, making them a primary target for activation sparsification.
  However, existing training-free methods suffer substantial model-quality degradation at high sparsity due to limitations in their channel-selection strategies.
  We observe that the SwiGLU intermediate state provides a highly effective channel-selection signal, but obtaining it requires costly dense computation.
  To address this, we present \emph{Prox}, a two-stage training-free framework for sparse SwiGLU FFNs.
  Prox hinges on the key insight: sparse execution requires only the channel mask induced by the intermediate state, which can be constructed from the magnitude ranking of its entries rather than their exact values.
  Specifically, Stage~1 uses input sparsity and quantized proxy weights to construct a shared mask; Stage~2 computes the selected channels exactly, enabling sparse execution of all three projections.
  Across ten LLMs from six model families, Prox outperforms training-free baselines at all sparsity levels, achieves up to a $1.99\times$ end-to-end decoding speedup at 70\% FFN sparsity, and is compatible with quantization and sparse attention.
\end{abstract}

\section{Introduction}

\begin{figure}[t]
  \centering
  \includegraphics[width=0.95\columnwidth]{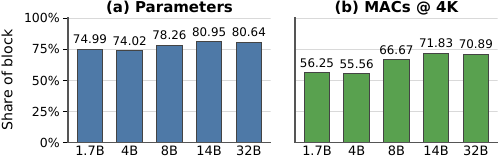}
  \caption{FFNs account for the majority of parameters and multiply-accumulate operations (MACs) under 4K context length in Qwen3 Transformer blocks across model scales. }
  \label{fig:qwen3_ffn_block_shares}
\end{figure}

Efficient inference is critical for deploying large language models (LLMs), especially in resource-constrained and latency-sensitive settings. In small-batch autoregressive decoding, parameter transfers from off-chip high-bandwidth memory (HBM) to on-chip storage dominate latency. Modern LLMs commonly adopt SwiGLU feed-forward networks (FFNs), whose three projection matrices account for most model parameters, memory traffic, and computation (Figure~\ref{fig:qwen3_ffn_block_shares}), making them a high-leverage target for acceleration.

Existing approaches accelerate LLM inference through weight quantization \citep{gpt3_int8,gptq,awq} and model pruning \citep{llm_pruner,sparsegpt,slicegpt,shortgpt,ig_pruning,sun2024simple}, which lower memory footprint or computational cost by altering model precision or structure.
Orthogonal to these static modifications, activation sparsity \citep{dejavu,cats,countdown,teal,sparsevit,raihan2020sparse,kurtz2020inducing,universal_properties_activation_sparsity} mitigates memory and compute bottlenecks by dynamically skipping low-salience channels and associated MACs. Prior works exploit input-dependent activation sparsity via either training-based predictors or training-free heuristics.
However, these predictors must be trained separately for each model, while existing training-free schemes often suffer substantial accuracy degradation as sparsity increases due to limitations in their channel-selection strategies.

Our analysis reveals two findings:
(1) The SwiGLU intermediate state provides a highly effective channel-selection signal but requires costly dense computation. Our oracle experiment demonstrates that selecting via this exact state incurs less than a 3\% relative perplexity increase at 70\% sparsity across most models.
(2) Sparse execution requires only the channel mask, which depends only on the magnitude ranking of the state entries rather than their exact values.
Our experiment shows that a low-cost input-sparse proxy preserves this ranking and closely matches the oracle mask.


Based on these two key findings, we propose \emph{Prox}, a two-stage training-free method for sparse SwiGLU inference. Stage~1 (\emph{lightweight proxy construction}) combines magnitude-based input sparsity with quantized proxy weights to estimate both SwiGLU branches and construct a shared intermediate-channel mask. Stage~2 (\emph{exact sparse computation}) computes the selected channels with the original up and gate weights, and applies the same mask to the down projection.
Prox further allocates computation between the two stages under a target sparsity budget and uses custom CUDA and Triton \citep{triton} kernels to reduce weight accesses and computation across all three projections.

We evaluate Prox on ten representative LLMs from six model families. The results show that Prox consistently outpaces state-of-the-art training-free baselines, particularly at 60-70\% FFN sparsity, yielding up to a $1.99\times$ end-to-end decoding speedup with minimal accuracy loss. Crucially, Prox is fully orthogonal to quantization and sparse attention, enabling synergistic efficiency gains.

Our main contributions are summarized as follows:
\begin{itemize}
  \item \textbf{Insight}. We identify the FFN intermediate state as an effective channel-selection signal and show that a low-cost proxy preserves the magnitude ranking needed to recover the mask induced by the exact state. This motivates proxy-based channel selection followed by exact computation of the retained channels.
  \item \textbf{Prox Framework}. We propose Prox, a training-free two-stage framework that uses quantized proxies for channel selection and computes retained values exactly, sparsifying all three SwiGLU projections while preventing error accumulation.
  \item \textbf{Comprehensive Evaluation}. Our extensive experiments demonstrate that Prox delivers superior accuracy-efficiency trade-offs across various models and sparsity levels, while maintaining full compatibility with quantization and sparse attention.
\end{itemize}

\section{Background and Related Work}

\subsection{SwiGLU Feed-Forward Networks}

LLMs are typically built with stacked Transformer layers, each comprising an attention block and an FFN. While early Transformer FFNs employed ReLU or GELU activation functions, Gated Linear Units (GLUs) \citep{dauphin2017language,shazeer2020glu} have largely superseded them.
Today, SwiGLU \citep{shazeer2020glu} remains the most widely adopted GLU variant in modern LLMs, with representative models including Qwen3 \citep{qwen3technicalreport}, Llama-3 \citep{llama3modelcard}, and Mistral \citep{jiang2023mistral7b}.

A SwiGLU FFN comprises three projection matrices: the up projection $W_{\mathrm{up}}\in\mathbf{R}^{d_{\mathrm{model}}\times d_{\mathrm{ff}}}$, the gate projection $W_{\mathrm{gate}}\in\mathbf{R}^{d_{\mathrm{model}}\times d_{\mathrm{ff}}}$, and the down projection $W_{\mathrm{down}}\in\mathbf{R}^{d_{\mathrm{ff}}\times d_{\mathrm{model}}}$. Given an input representation $\mathbf{x}\in\mathbf{R}^{d_{\mathrm{model}}}$, its forward computation is defined as
\[
  \mathrm{FFN}(\mathbf{x})
  =
  \bigl(\mathbf{x}W_{\mathrm{up}}\odot \mathrm{SiLU}(\mathbf{x}W_{\mathrm{gate}})\bigr)
  W_{\mathrm{down}}.
\]
For subsequent discussion, we define three variables:
\[
  \mathbf{u}=\mathbf{x}W_{\mathrm{up}},
  \quad
  \mathbf{h}=\mathrm{SiLU}(\mathbf{x}W_{\mathrm{gate}}),
  \quad
  \mathbf{s}=\mathbf{u}\odot\mathbf{h},
\]
denoting the up-branch activation, the gated nonlinear activation, and the intermediate SwiGLU state, respectively.

\subsection{Activation Sparsity}

Activation sparsity mitigates both HBM-to-register memory bandwidth overhead and redundant computations on low-salience channels.
For a linear projection
$
\mathbf{y}=\mathbf{x}W
$
with $\mathbf{x}\in\mathbf{R}^{d_{\mathrm{in}}}$ and $W\in\mathbf{R}^{d_{\mathrm{in}}\times d_{\mathrm{out}}}$, sparsity can be enforced along either the input dimension or output dimension.
Let $\mathcal{I}(\mathbf{m})=\{i\mid m_i=1\}$ denote the set of indices retained by a binary mask $\mathbf{m}$ along the target dimension.


\textbf{Input sparsity} applies a mask $\mathbf{m}\in\{0,1\}^{d_{\mathrm{in}}}$ to the input dimension. Specifically, $m_i=0$ masks the $i$-th input channel, eliminating the multiplication of $x_i$ with the corresponding row of $W$.
\[
  \Pi^{\mathrm{in}}(\mathbf{x}, W, \mathbf{m})
  =
  \mathbf{x}_{[\mathcal{I}(\mathbf{m})]}W_{[\mathcal{I}(\mathbf{m}),:]}.
\]

\textbf{Output sparsity} applies a mask $\mathbf{m}\in\{0,1\}^{d_{\mathrm{out}}}$ to the output dimension and computes only the selected output coordinates:
\[
  \Pi^{\mathrm{out}}(\mathbf{x}, W, \mathbf{m})
  =
  \mathbf{x}W_{[:,\mathcal{I}(\mathbf{m})]},
\]
where the selected columns of $W$ yield the retained output channels.

\begin{figure*}[t]
  \centering
  \includegraphics[width=0.9\textwidth]{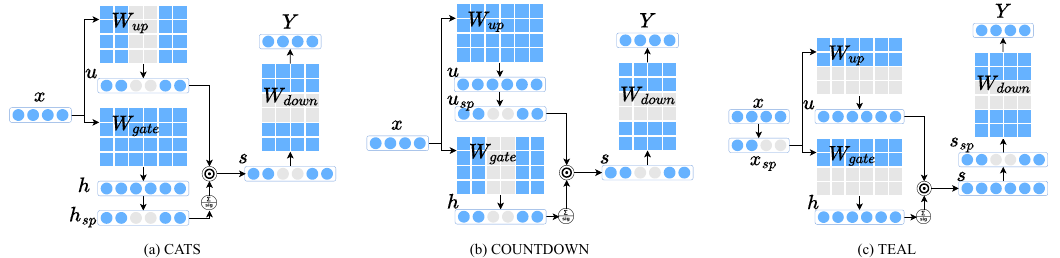}
  \caption{Computation patterns of representative sparsification methods for SwiGLU FFNs.}
  \label{fig:baseline_computation}
\end{figure*}

The evolution of activation sparsification closely tracks the architectural transition of Transformer FFNs from ReLU-based designs to modern SwiGLU variants.

Early activation sparsification methods primarily target ReLU-based FFNs, skipping input-dependent inactive or low-magnitude neurons to reduce computation and memory traffic \citep{lazy_neuron,dejavu,llm_in_a_flash}.
While these approaches excel with ReLU-induced exact or near-exact zeros, they cannot directly adapt to SiLU-based SwiGLU FFNs.

A second line of work, ReLUfication, converts non-ReLU FFNs to ReLU-style variants via activation replacement and post-training adaptation \citep{relu_strikes_back,relu2_wins,prosparse,turbo_sparse}. However, these approaches typically require explicit architectural modifications or resource-intensive retraining, such as continued training or knowledge distillation.

For native SwiGLU FFNs, predictor-based methods estimate salient intermediate channels before executing the full FFN. For instance, \citet{countdown} require training model-specific, layer-wise low-rank predictors on FFN inputs to select channels from the intermediate state $\mathbf{s}$. However, limited prediction accuracy can degrade downstream model quality.

In contrast, training-free methods sparsify native SwiGLU FFNs using activation magnitudes. They differ mainly in two dimensions: which activation signals guide channel selection and how the resulting masks induce sparse execution. Figure~\ref{fig:baseline_computation} compares three representative approaches along these two dimensions. CATS \citep{cats} selects intermediate channels based on the magnitude of the densely computed gate activation $\mathbf{h}$, inducing output sparsity in the up projection and input sparsity in the down projection (Figure~\ref{fig:baseline_computation}(a)). COUNTDOWN \citep{countdown} instead performs channel selection using the magnitude of the densely computed up-branch activation $\mathbf{u}$, yielding output sparsity in the gate projection and input sparsity in the down projection (Figure~\ref{fig:baseline_computation}(b)).\footnote{Throughout this paper, COUNTDOWN refers to the training-free M-COUNTDOWN variant unless stated otherwise.}
Unlike these single-branch methods, TEAL \citep{teal} applies input sparsity to $\mathbf{x}$ in the up and gate projections and to $\mathbf{s}$ in the down projection (Figure~\ref{fig:baseline_computation}(c)) simultaneously. R-Sparse follows a similar input-sparsity pattern while incorporating low-rank compensation across linear layers \citep{zhang2025rsparse}.

Across the three representative methods in Figure~\ref{fig:baseline_computation}, model quality deteriorates substantially as sparsity increases, exposing key limitations in their channel-selection strategies.
CATS and COUNTDOWN estimate intermediate-channel salience solely from $\mathbf{h}$ and $\mathbf{u}$, respectively; each relies on a single-branch signal that captures only partial information of the joint intermediate state $\mathbf{s}=\mathbf{u}\odot\mathbf{h}$, leading to misranked salient channels.
TEAL instead applies input sparsity at two successive points: first to $\mathbf{x}$, the shared input to the up and gate projections, and then to the resulting intermediate state $\mathbf{s}$, the input to the down projection.
The latter therefore operates on an already approximated intermediate state, allowing errors from the two sparsification steps to compound through the FFN.


These limitations motivate a more effective channel-selection mechanism tailored for training-free SwiGLU sparsification.

\begin{figure*}[t]
  \centering
  \includegraphics[width=\textwidth]{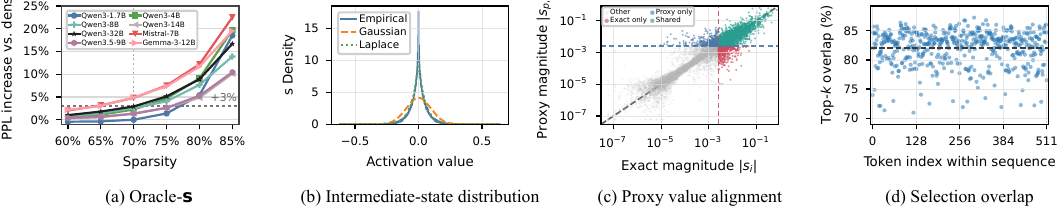}
  \caption{(a) Perplexity increase vs. sparsity under oracle-$\mathbf{s}$ channel selection. (b) Empirical distribution of $\mathbf{s}$ in Qwen3-8B Layer~8 and fitted Gaussian and Laplace densities; (c) Exact vs. proxy intermediate-state magnitudes for one representative token from Qwen3-8B Layer~3 at 70\% effective sparsity; colors indicate whether each channel is retained by both top-ranking selections, only one selection, or neither. The two selections achieve 82.77\% overlap. (d) Exact/proxy top-ranking overlap across 512 tokens; the dashed line marks the mean overlap of 82.04\%.}
  \label{fig:oracle_s_ppl_vs_sparsity}
\end{figure*}

\section{Motivation}

\subsection{Channel Selection with Intermediate State $\mathbf{s}$}
Following the weighted-sum view of SwiGLU FFNs \citep{countdown}, the FFN output is a weighted sum of down-projection rows:
\[
  \mathrm{FFN}(\mathbf{x})=\mathbf{s}W_{\mathrm{down}}
  =\sum_{i=1}^{d_{\mathrm{ff}}} s_i W_{\mathrm{down}}[i,:].
\]

\textbf{Analysis.} The input-dependent coefficient $s_i$ governs the output of each row, making $|s_i|$ a natural metric for channel selection.
Beyond measuring output contribution, the intermediate state
$\mathbf{s}$ is natively aligned with the shared intermediate dimension of all three FFN projections: each index corresponds to an output channel in both up- and gate-projections, as well as the matching input channel in the down-projection.
Accordingly, if the exact intermediate state $\mathbf{s}$, denoted as $\mathbf{s}_{\mathrm{exact}}$, were available prior to FFN execution, a single $\mathbf{s}$-based mask would induce output sparsity in the up- and gate-projections alongside input sparsity in the down-projection, thereby sparsifying all three projections in a unified and consistent manner.

\textbf{Assumption and oracle evaluation.}
To assess the performance upper bound of $\mathbf{s}$-based channel selection on modern SwiGLU models, we evaluate an ideal diagnostic baseline, \emph{oracle-$\mathbf{s}$}, under the assumption that $\mathbf{s}_{\mathrm{exact}}$ is available at no cost prior to FFN execution.
Given a target sparsity $\rho$, oracle-$\mathbf{s}$ retains the $(1-\rho)d_{\mathrm{ff}}$ coordinates of $\mathbf{s}$ with the largest magnitudes to construct a mask across all intermediate channels.

Figure~\ref{fig:oracle_s_ppl_vs_sparsity}(a) empirically validates the above analysis: \emph{oracle-$\mathbf{s}$} preserves model quality across moderate-to-high sparsity levels. Even at 70\% sparsity, most tested models experience less than a 3\% relative perplexity increase over their dense counterparts. Figure~\ref{fig:oracle_s_ppl_vs_sparsity}(b) provides a complementary explanation for this robustness. The distribution of $\mathbf{s}$ is sharply peaked and Laplacian-like around zero, indicating that many channels carry negligible magnitudes and can be safely skipped with limited impact on FFN output.

Consequently, while the channel-selection signal provided by $\mathbf{s}_{\mathrm{exact}}$ serves as an ideal reference, the key challenge in practice is that the exact intermediate state is unavailable until after executing the dense computations for the up- and gate-projections.

\subsection{Low-Cost Input-Sparse Proxy}


In fact, sparse execution does not require the exact values of $\mathbf{s}$ in advance; it only requires the binary channel mask that $\mathbf{s}_{\mathrm{exact}}$ would induce. Since this mask is determined by the magnitude ranking of entries in $\mathbf{s}$, our key insight is that an effective proxy needs only to preserve relative channel rankings rather than accurately construct the exact values of $\mathbf{s}$.
We therefore design a low-cost proxy that bypasses full projection computations and serves exclusively to derive the channel mask.

Magnitude-based input sparsity is naturally suited for preserving this ranking, as it minimizes perturbations to the projected values that determine the channel ranking. Let $\mathbf{x}_{\mathrm{sp}}$ denote the sparse vector retaining the largest-magnitude entries of $\mathbf{x}$, and let $\mathbf{r}=\mathbf{x}-\mathbf{x}_{\mathrm{sp}}$ be the discarded residual.
For a fixed density, this selection minimizes $\|\mathbf{r}\|_2$ across all coordinate-sparse approximations, thus discarding the minimum amount of input energy. For any projection matrix $W$, this residual directly bounds the resulting output error:
\[
  \begin{array}{r@{}l}
    \|\mathbf{x}W-\mathbf{x}_{\mathrm{sp}}W\|_2
    &=\|\mathbf{r}W\|_2
    \leq \|\mathbf{r}\|_2\|W\|_2.
  \end{array}
\]
Input sparsity thus reduces the effective input dimension of the projection, bounds the overall perturbation, and preserves all output channels.


Applying the same sparse input to the up and gate projections yields proxy activations $\tilde{\mathbf{u}}$ and $\tilde{\mathbf{h}}$, whose element-wise product $\tilde{\mathbf{s}}=\tilde{\mathbf{u}}\odot\tilde{\mathbf{h}}$ is used to rank intermediate channels by magnitude.
Crucially, the proxy activations are used exclusively to construct the channel mask and are never propagated downstream through the FFN. Their approximation errors affect model outputs solely via channel-selection mismatches, which occur primarily when perturbations shift near-threshold channels across the selection boundary.

As Figure~\ref{fig:oracle_s_ppl_vs_sparsity}(c) illustrates, while exact and input-sparse proxy values exhibit numerical differences, they yield largely overlapping channel sets. Figure~\ref{fig:oracle_s_ppl_vs_sparsity}(d) further corroborates strong overlap among top-ranked channels across tokens. This ranking stability motivates our two-stage design: the input-sparse proxy is used exclusively to construct the channel mask, followed by exact computation of the retained channels.

Appendix~\ref{app:activation_distributions} provides additional quantitative analyses of intermediate-state compressibility and proxy ranking preservation across representative FFN layers.

\section{Methodology}

\subsection{Two-Stage Sparse Inference}
\begin{figure}[!t]
  \centering
  \includegraphics[width=0.7\columnwidth]{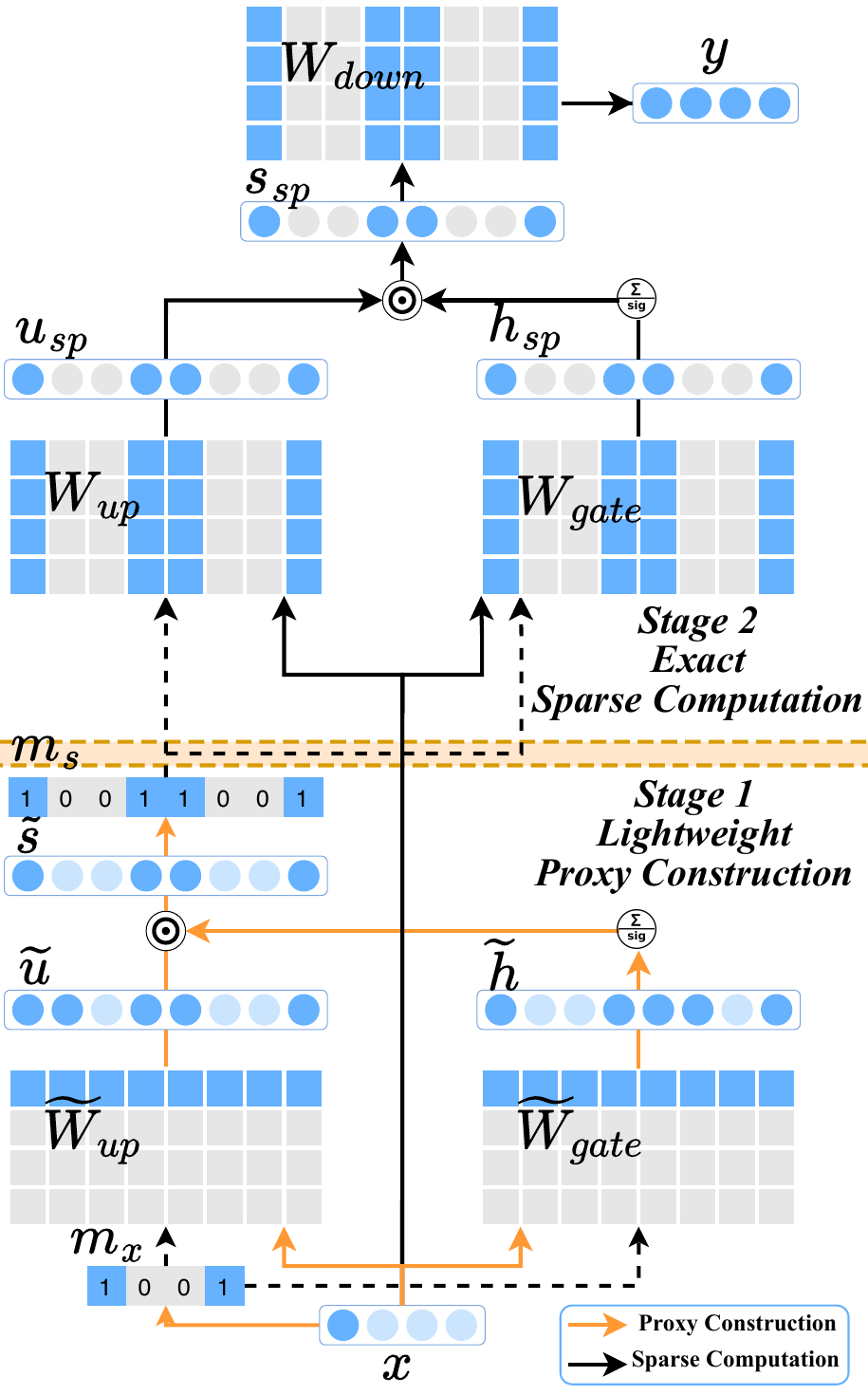}
  \caption{Overview of Prox. A quantized proxy derives a shared intermediate-channel mask; then the selected intermediate channels are computed with the original weights.}
  \label{fig:method_v2}
\end{figure}

We propose Prox, a two-stage training-free framework for sparse SwiGLU FFN inference. As illustrated in Figure~\ref{fig:method_v2}, given an FFN input $\mathbf{x}$, Prox proceeds in the following two stages.

\textbf{Stage 1: lightweight proxy construction.}
The proxy activations are used exclusively to construct the channel mask $\mathbf{m}_{\mathbf{s}}$. Their approximation errors thus affect outputs only by altering channel selection, rather than propagating as activation values. This error isolation enables more aggressive approximation in the proxy path: Prox combines input sparsity with quantized proxy weights to reduce both the effective input dimension and weight memory overhead.




For FFN layer $\ell$, let $s_1$ and $s_2$ denote the target sparsities of Stage~1 input entries and Stage~2 intermediate channels, respectively. To achieve the target input sparsity $s_1$, Prox constructs a binary input mask $\mathbf{m}_{\mathbf{x}}\in\{0,1\}^{d_{\mathrm{model}}}$ by retaining entries with magnitudes exceeding a layer-wise threshold:
\[
  [\mathbf{m}_{\mathbf{x}}]_i
  =\mathbf{1}\!\left[|x_i|\geq\tau_{\mathbf{x},\ell}^{s_1}\right],
  \qquad i=1,\ldots,d_{\mathrm{model}},
\]
where $\mathbf{1}[\cdot]$ denotes the indicator function.
The threshold $\tau_{\mathbf{x},\ell}^{s_1}$ is calibrated offline on layer-$\ell$ inputs to approximately match the target sparsity $s_1$ over a calibration dataset.
Given quantized proxy weights $\widetilde{W}_{\mathrm{up}}$ and $\widetilde{W}_{\mathrm{gate}}$, the proxy path leverages this input sparsity to compute only entries retained by $\mathbf{m}_{\mathbf{x}}$:
\[
  \tilde{\mathbf{u}}
  =\Pi^{\mathrm{in}}(\mathbf{x},\widetilde{W}_{\mathrm{up}},\mathbf{m}_{\mathbf{x}}),
  \;
  \tilde{\mathbf{h}}
  =\mathrm{SiLU}\!\left(\Pi^{\mathrm{in}}(\mathbf{x},\widetilde{W}_{\mathrm{gate}},\mathbf{m}_{\mathbf{x}})\right).
\]
It then constructs the proxy intermediate signal
$
\tilde{\mathbf{s}}=\tilde{\mathbf{u}}\odot\tilde{\mathbf{h}}.
$

For the target intermediate-channel sparsity $s_2$, Prox applies a corresponding layer-wise magnitude threshold to the proxy intermediate state:
\[
  [\mathbf{m}_{\mathbf{s}}]_j
  =\mathbf{1}\!\left[|\tilde{s}_j|\geq\tau_{\mathbf{s},\ell}^{s_2}\right],
  \qquad j=1,\ldots,d_{\mathrm{ff}},
\]
where $\mathbf{m}_{\mathbf{s}}\in\{0,1\}^{d_{\mathrm{ff}}}$ denotes the active channels retained for the exact computation path.
Analogous to $\tau_{\mathbf{x},\ell}^{s_1}$, the threshold $\tau_{\mathbf{s},\ell}^{s_2}$ is calibrated offline and remains fixed during inference.

\textbf{Stage 2: exact sparse computation.}
Using the channel mask $\mathbf{m}_{\mathbf{s}}$ from Stage~1, Prox performs exact SwiGLU computations exclusively over the selected channels using the original full-precision model weights.
\[
  \begin{array}{r@{}l@{\,}r@{}l}
    \mathbf{u}_{\mathrm{sp}}\!
    &=\!\Pi^{\mathrm{out}}(\mathbf{x},\mkern-4mu W_{\mathrm{up}},\mkern-4mu\mathbf{m}_{\mathbf{s}}),
    &\mathbf{h}_{\mathrm{sp}}\!
    &=\!\mathrm{SiLU}\!\left(\Pi^{\mathrm{out}}(\mathbf{x},\mkern-4mu W_{\mathrm{gate}},\mkern-4mu\mathbf{m}_{\mathbf{s}})\right),\\
    \mathbf{s}_{\mathrm{sp}}\!
    &=\!\mathbf{u}_{\mathrm{sp}}\odot\mathbf{h}_{\mathrm{sp}},
    &\mathbf{y}\!
    &=\!\Pi^{\mathrm{in}}(\mathbf{s}_{\mathrm{sp}},\!W_{\mathrm{down}},\!\mathbf{m}_{\mathbf{s}}).
  \end{array}
\]

\subsection{Sparsity Allocation between Two Stages}
\label{sec:sparsity_tradeoff}
$s_1$ and $s_2$ jointly govern the overall computational cost of Prox. Reducing Stage~1 sparsity retains more input entries to yield a more informative proxy signal, while reducing Stage~2 sparsity preserves more intermediate channels for exact SwiGLU computation.
Since both adjustments increase computation, $s_1$ and $s_2$ must be jointly optimized under a target FFN compute budget.

We model computational overhead with a normalized cost framework.
Dense \textit{up-}, \textit{gate-}, and \textit{down-}projections each carry a unit cost, yielding a total dense FFN computational cost $C_{\mathrm{Dense}}=3$.
Let $\alpha$ denote the normalized cost of a quantized proxy projection relative to a full dense projection.
Given sparsity $s_1$ and $s_2$, Stage~1 executes two proxy projections over a $(1-s_1)$ fraction of the input dimension, while Stage~2 executes three full-precision projections over a $(1-s_2)$ fraction of the intermediate channels.
The total normalized cost is:
\[
  C_{\mathrm{Prox}}=2\alpha(1-s_1)+3(1-s_2).
\]
We define the resulting effective sparsity $e$ relative to the dense FFN baseline as:
\[
  e=1-\frac{C_{\mathrm{Prox}}}{C_{\mathrm{Dense}}}
  =s_2-\frac{2\alpha(1-s_1)}{3}.
\]



Our budget allocation policy is inspired by the findings of oracle-$\mathbf{s}$ in Figure~\ref{fig:oracle_s_ppl_vs_sparsity}(a). The results demonstrate that sparsifying approximately 70\% of the exact intermediate channels incurs negligible performance degradation across the evaluated models, highlighting substantial channel redundancy in Stage~2 exact computation. We thus set $s_{\mathrm{ref}}=0.7$ as a quality-preserving anchor for Stage~2. Given a target effective sparsity $e$, we initially set $s_2=s_{\mathrm{ref}}$ and allocate the remaining compute budget to Stage~1 according to the effective-sparsity relation above.

However, Stage~1 sparsity cannot be increased indefinitely; removing excessive input coordinates diminishes the fidelity of the proxy signal, thereby compromising the reliability of channel selection.
To mitigate this, we constrain $s_1$ to $[s_{1,\min},s_{1,\max}]=[0,0.7]$, ensuring at least 30\% of the input entries are retained.

If the Stage~1 sparsity implied by $s_2=s_{\mathrm{ref}}$ falls outside this interval, we clamp $s_1$ to the nearest bound and dynamically adjust $s_2$ to satisfy the target effective sparsity $e$. Detailed budget allocations across operating points are provided in Appendix~\ref{app:stagewise_configs}.


\subsection{Hardware-Aware Implementation}

To realize practical acceleration, we construct the Stage~1 proxy weights via symmetric per-row INT4 quantization of the original FP16 $W_{\mathrm{up}}$ and $W_{\mathrm{gate}}$ matrices, keeping the compact proxies resident in GPU memory. We implement dedicated CUDA and Triton kernels tailored for single-batch autoregressive decoding.

Stage~1 employs a fused split-$N$ CUDA kernel that evaluates the up- and gate-proxy projections together, sharing hidden-state reads and skipping input coordinates rejected by the calibrated threshold. INT4 unpacking is performed directly in registers, while per-thread partial sums are reduced in FP32. The resulting projection sums are rescaled and combined through SwiGLU to form the proxy intermediate state $\tilde{\mathbf{s}}$, whose magnitude is thresholded to produce the Stage~2 channel mask $\mathbf{m}_{\mathbf{s}}$ without storing separate FP16 up and gate proxy vectors.
Launch geometries and reduction strategies are autotuned across GPU architectures, matrix shape, and sparsity regimes.

In Stage~2, a fused output-sparse kernel computes the exact up and gate projections together with SwiGLU exclusively over active channels selected by $\mathbf{m}_{\mathbf{s}}$; the down projection then executes an input-sparse GEMV over these channels. Mask handling remains inside the main computation loops, restricting weight accesses and MACs to active coordinates and channels.




\section{Experiments}

\begin{table*}[t]
  \centering
  \footnotesize
  \setlength{\tabcolsep}{0pt}
  \renewcommand{\arraystretch}{1.08}
  \begin{tabular}{@{}L{0.135\textwidth}C{0.095\textwidth}*{10}{C{0.0765\textwidth}}@{}}
    \toprule
    & & \multicolumn{4}{c}{\textbf{Qwen3}}
    & \multicolumn{2}{c}{\textbf{Qwen3.5}}
    & \multicolumn{1}{C{0.0765\textwidth}}{\textbf{\shortstack{Ministral}}}
    & \multicolumn{1}{C{0.0765\textwidth}}{\textbf{Mistral}}
    & \multicolumn{1}{C{0.0765\textwidth}}{\textbf{Llama3}}
    & \multicolumn{1}{C{0.0765\textwidth}}{\textbf{Gemma3}} \\
    \cmidrule(lr){3-6}
    \cmidrule(lr){7-8}
    \cmidrule(lr){9-9}
    \cmidrule(lr){10-10}
    \cmidrule(lr){11-11}
    \cmidrule(lr){12-12}
    \textbf{Variant} & \textbf{Sparsity} & \textbf{1.7B} & \textbf{4B} & \textbf{8B} & \textbf{14B} & \textbf{4B} & \textbf{9B} & \textbf{3.3B} & \textbf{7B} & \textbf{8B} & \textbf{12B} \\
    \midrule
    Dense & 0 & 61.4 & 71.1 & 76.1 & 77.9 & 69.3 & 70.9 & 69.9 & 71.1 & 70.1 & 77.9 \\
    \midrule
    CATS & 40\% & 35.3 & 44.9 & 59.6 & 67.9 & 59.1 & 61.8 & 57.7 & 63.7 & 65.9 & 76.4 \\
    COUNTDOWN & 40\% & 59.3 & 69.4 & 73.6 & 75.7 & 63.0 & 67.0 & 65.6 & 66.8 & 68.7 & 77.1 \\
    TEAL & 40\% & 59.4 & \textbf{71.0} & \textbf{76.1} & 77.3 & 67.3 & 70.0 & \textbf{69.3} & 68.0 & \textbf{69.8} & 75.2 \\
    Prox & 40\% & \textbf{61.5} & 70.8 & 75.7 & \textbf{77.6} & \textbf{69.6} & \textbf{70.4} & 68.7 & \textbf{69.3} & 69.7 & \textbf{77.6} \\
    \midrule
    CATS & 50\% & 28.5 & 33.3 & 49.9 & 53.0 & 51.9 & 57.0 & 36.4 & 46.7 & 44.1 & 62.8 \\
    COUNTDOWN & 50\% & 51.2 & 65.9 & 71.4 & 75.2 & 60.6 & 62.6 & 64.0 & 62.2 & 61.3 & 73.2 \\
    TEAL & 50\% & 58.3 & 67.9 & 73.4 & \textbf{77.0} & 66.7 & \textbf{69.6} & 67.2 & 67.6 & 68.3 & 70.6 \\
    Prox & 50\% & \textbf{59.5} & \textbf{68.8} & \textbf{74.6} & 76.6 & \textbf{66.8} & 69.5 & \textbf{68.1} & \textbf{67.7} & \textbf{70.1} & \textbf{77.5} \\
    \midrule
    CATS & 60\% & 27.3 & 26.3 & 27.6 & 31.9 & 36.3 & 43.5 & 27.6 & 26.7 & 26.7 & 32.9 \\
    COUNTDOWN & 60\% & 33.7 & 45.0 & 56.8 & 66.7 & 49.9 & 55.1 & 45.9 & 50.0 & 44.2 & 54.0 \\
    TEAL & 60\% & 55.9 & 66.2 & 71.4 & 75.6 & 64.6 & 68.2 & 63.4 & 64.6 & 66.8 & 66.8 \\
    Prox & 60\% & \textbf{59.3} & \textbf{66.4} & \textbf{73.3} & \textbf{76.1} & \textbf{65.0} & \textbf{68.4} & \textbf{67.7} & \textbf{67.0} & \textbf{67.2} & \textbf{77.5} \\
    \midrule
    TEAL & 70\% & 41.8 & 52.8 & 63.3 & 70.3 & 54.1 & 61.2 & 52.2 & 59.4 & 57.1 & 62.0 \\
    Prox & 70\% & \textbf{56.5} & \textbf{58.7} & \textbf{68.6} & \textbf{74.8} & \textbf{59.1} & \textbf{63.0} & \textbf{65.3} & \textbf{67.5} & \textbf{63.4} & \textbf{74.8} \\
    \bottomrule
  \end{tabular}
  \caption{Aggregate downstream task scores across model families and sparsity levels. Higher is better.}
  \label{tab:downstream}
\end{table*}

\textbf{Models and datasets.}
We evaluate Prox on SwiGLU-based model families, including Qwen3 \citep{qwen3technicalreport}, Qwen3.5 \citep{qwen3.5}, Ministral, Mistral \citep{jiang2023mistral7b}, and Llama-3 \citep{llama3modelcard}.
To assess generalization beyond SwiGLU, we also evaluate Gemma-3 \citep{gemma_2025} that features a GeGLU-style FFN.
Downstream accuracy is evaluated via the EleutherAI LM Harness \citep{eval-harness} across six tasks: 8-shot GSM8K \citep{cobbe2021training}, 5-shot MMLU \citep{mmlu}, 10-shot HellaSwag \citep{zellers2019hellaswag}, zero-shot ARC-Easy and 25-shot ARC-Challenge \citep{arc}, and zero-shot PIQA \citep{bisk2020piqa}.


\textbf{Baselines.}
We compare Prox against three training-free activation sparsity baselines: CATS \citep{cats}, COUNTDOWN \citep{countdown}, and TEAL \citep{teal}.
We use M-COUNTDOWN as the COUNTDOWN baseline because it achieves higher average downstream scores than D-COUNTDOWN under practical inference. 

To ensure a fair comparison focused on FFN sparsification, we apply all methods exclusively to FFN projections and report sparsity at the FFN-module level.
Specifically, for CATS and COUNTDOWN, a projection-level sparsity of $k$ yields an effective sparsity of $2k/3$; TEAL's effective sparsity is averaged across its three FFN projections, while Prox's effective sparsity follows the cost model in Sec.~\ref{sec:sparsity_tradeoff}.
Although the theoretical INT4-to-FP16 bit-width ratio implies $\alpha=0.25$, it ignores quantization overhead. We therefore set $\alpha=1/3$, a single rounded value within the measured INT4-to-FP16 GEMV latency-ratio range across three GPUs and four Qwen3 scales (details in Appendix~\ref{app:alpha_validation}).

These baselines primarily target decoding-phase sparsification.
Nevertheless, for log-likelihood tasks, we sparsify the full sequence to ensure hidden states are computed under sparsity; for generation tasks such as GSM8K, we maintain a dense prompt prefill and sparsify decoding exclusively.

\subsection{Results}

\textbf{Downstream accuracy.}
As shown in Table~\ref{tab:downstream}, Prox demonstrates its superiority at high sparsity levels while remaining competitive with TEAL at 40\% sparsity. Specifically, Prox surpasses TEAL on 8 out of 10 evaluated models at 50\% sparsity and consistently outperforms it across all 10 models at 60\% and 70\% sparsity. In contrast, CATS and COUNTDOWN exhibit premature degradation as sparsity increases; restricted to sparsifying only two FFN matrices, they fail to reach the 70\% operating point sustained by TEAL and Prox. Detailed perplexity results are provided in Appendix~\ref{app:ppl_results}.


\textbf{Role of exact computation.}
CATS and COUNTDOWN approximate the SwiGLU intermediate state  $\mathbf{s}$ via a single branch, rendering channel selection unreliable under high sparsity.
TEAL can be viewed as a simplified Prox since it directly reuses a sparse approximation of $\mathbf{s}$ for FFN computation; while this reduces computation, it propagates proxy errors into the output. In contrast, Prox uses the proxy only for channel selection and computes the selected values exactly, improving robustness at high sparsity.

\begin{figure}[t]
  \centering
  \includegraphics[width=\columnwidth]{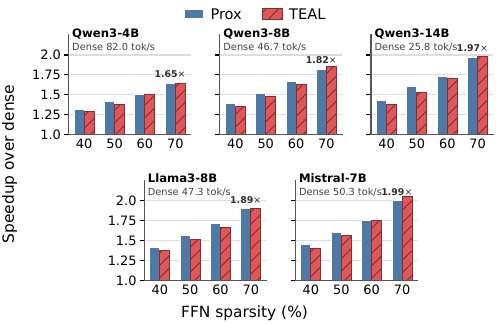}
  \caption{Single-batch decoding speedup on an NVIDIA A6000 GPU. Headers give dense throughput (tokens/s), and numeric annotations indicate Prox's speedup at 70\% sparsity.}
  \label{fig:e2e_speedup}
\end{figure}

\textbf{End-to-end decoding speedup.}
We benchmark Prox and TEAL under single-batch end-to-end decoding on an NVIDIA A6000 GPU. As Figure~\ref{fig:e2e_speedup} shows, both methods effectively leverage FFN sparsity to boost decoding throughput across models. Prox achieves a $1.51$--$1.99\times$ speedup at 60--70\% sparsity.
At 70\% sparsity, Prox differs from TEAL in throughput by no more than 2.9\% across all evaluated models and improves the average downstream task performance by 14.4\%.
A detailed latency breakdown and additional throughput measurements on A100 and RTX 3090 GPUs are provided in Appendix~\ref{app:e2e_3090}.

\begin{figure}[t]
  \centering
  \includegraphics[width=0.85\columnwidth]{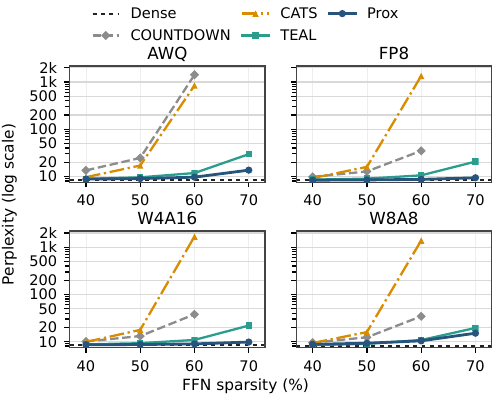}
  \caption{Perplexity of quantized Qwen3-8B across FFN sparsity levels (log scale; lower is better).}
  \label{fig:quantized_ppl}
\end{figure}

\textbf{Compatibility with model quantization.}
Model quantization and FFN sparsification are complementary techniques that can be combined for efficient inference. We evaluate Prox and three baselines on Qwen3-8B across four quantization configurations (AWQ, FP8, W4A16, W8A8) by measuring WikiText perplexity. As Figure~\ref{fig:quantized_ppl} shows, Prox consistently achieves the lowest perplexity across all configurations and sparsities, with its performance advantage becoming increasingly pronounced in high-sparsity regimes. Combining low-bit weights with activation sparsity can further reduce weight traffic and improve inference speed, but realizing these gains requires specialized kernels that jointly support sparse execution and low-bit arithmetic, which we leave to future work.

\begin{table}[t]
  \centering
  \footnotesize
  \setlength{\tabcolsep}{1pt}
  \renewcommand{\arraystretch}{1.12}
  \newcommand{\attnspeed}[2]{#1 ($#2\times$)}
  \begin{tabular*}{\columnwidth}{@{\extracolsep{\fill}}lccc@{}}
    \toprule
    & \multicolumn{2}{c}{\textbf{Decoding Throughput (tok/s)}} & \textbf{Accuracy} \\
    \cmidrule(lr){2-3}
    \cmidrule(lr){4-4}
    \textbf{Variant} & \textbf{16K} & \textbf{32K} & \textbf{\shortstack{LongBench /\\RULER}} \\
    \midrule
    Dense (Qwen3-8B) & \attnspeed{33.1}{1.00} & \attnspeed{23.9}{1.00} & 100.0 / 100.0 \\
    \midrule
    RocketKV & \attnspeed{40.7}{1.23} & \attnspeed{41.0}{1.72} & 100.0 / 100.0 \\
    RocketKV + Prox & \attnspeed{63.6}{1.92} & \attnspeed{62.5}{2.62} & 96.6 / 98.0 \\
    \midrule
    HShare & \attnspeed{34.6}{1.05} & \attnspeed{39.0}{1.63} & 93.5 / 99.0 \\
    HShare + Prox & \attnspeed{40.0}{1.21} & \attnspeed{42.1}{1.76} & 92.5 / 100.0 \\
    \bottomrule
  \end{tabular*}
  \caption{Compatibility of Prox (60\% FFN sparsity) with different sparse-attention backends.}
  \label{tab:sparse_attention_speed}
\end{table}

\textbf{Compatibility with sparse attention.}
Prox operates solely on FFNs and is thus orthogonal to sparse-attention techniques, enabling combined computational savings across both attention and FFN modules. Table~\ref{tab:sparse_attention_speed} presents results integrated with RocketKV \citep{rocketkv} and HShare \citep{hshare} at 16K and 32K sequence lengths.
Integrating Prox with either backend further improves decoding throughput while largely preserving long-context accuracy.


\subsection{Ablation Studies}

\begin{table}[t]
  \centering
  \footnotesize
  \setlength{\tabcolsep}{1.6pt}
  \renewcommand{\arraystretch}{1.14}
  \begin{tabular*}{\columnwidth}{@{\extracolsep{\fill}}lcccccccc@{}}
    \toprule
    & \multicolumn{4}{c}{\textbf{Qwen3-8B}} & \multicolumn{4}{c}{\textbf{Qwen3-14B}} \\
    \cmidrule(lr){2-5}
    \cmidrule(lr){6-9}
    \textbf{Variant} & \textbf{40\%} & \textbf{50\%} & \textbf{60\%} & \textbf{70\%}
    & \textbf{40\%} & \textbf{50\%} & \textbf{60\%} & \textbf{70\%} \\
    \midrule
    \textbf{Prox} & 75.7 & 74.6 & 73.3 & 68.6 & 77.6 & 76.6 & 76.1 & 74.8 \\
    \midrule
    \textbf{A1} & 72.5 & 71.6 & 67.3 & 59.6 & 76.4 & 76.0 & 74.9 & 70.1 \\
    \textbf{A2} & 74.4 & 74.3 & 68.9 & 65.7 & 77.3 & 76.4 & 75.9 & 74.6 \\
    \textbf{A3} & 72.6 & 69.8 & 67.2 & 44.3 & 76.9 & 75.5 & 71.9 & 56.7 \\
    \bottomrule
  \end{tabular*}
  \caption{Ablation results on aggregate downstream scores.}
  \label{tab:ablation}
\end{table}

\textbf{A1: Effect of proxy-weight precision.}
To evaluate whether higher precision improves channel selection, A1 replaces the INT4 proxy weights with FP16 weights, reallocating the two-stage compute under the same target budget to accommodate the increased proxy cost. As Table~\ref{tab:ablation} shows, A1 consistently underperforms Prox across all eight settings, with performance deficits reaching 9.0 and 4.7 points at 70\% sparsity on Qwen3-8B and Qwen3-14B, respectively. This confirms that higher proxy precision does not lead to better overall quality under fixed budgets, justifying our low-bit proxy design in Stage 1.


\textbf{A2: Effectiveness of target-aware compute allocation.}
A2 fixes Stage~1 sparsity at $s_1 = 0.8$ across all targets and adjusts Stage~2 to meet the budget, whereas Prox dynamically adapts both $s_1$ and $s_2$ to the target sparsity. We select $s_1 = 0.8$ as a representative low-cost fixed baseline (a full Qwen3-8B sweep over $s_1$ is provided in Appendix~\ref{app:a2_sweep}). Across all eight model--sparsity settings, target-aware allocation matches or outperforms this fixed baseline, yielding gains of up to 4.4 points on Qwen3-8B and 0.2--0.3 points on Qwen3-14B. Together with the full sweep, these results demonstrate the necessity of dynamically adapting stage-wise allocations across operating points.


\textbf{A3: Necessity of exact computation.}
A3 retains Prox's complete setup but skips Stage 2 exact computation, directly feeding selected proxy values into the down projection. This evaluates whether quantized proxies can serve as final activation values rather than just channel selectors. Prox outperforms A3 across all settings, with performance gaps reaching 24.3 points on Qwen3-8B and 18.1 points on Qwen3-14B at 70\% sparsity. This demonstrates that while quantized proxies effectively identify salient channels, they cannot replace exact activations, confirming that Stage~2 exact computation is essential.


\subsection{Discussion}
Prox currently targets single-batch autoregressive decoding, a typical setting for latency-sensitive, on-device LLM applications. Consequently, its primary limitation is the lack of support for large-batch serving.
Extending proxy-guided sparsification and its underlying sparse kernels to efficient batched inference remains a promising avenue for future exploration.
Additionally, retaining Stage~1 INT4 proxy weights in GPU memory introduces an approximate 12\% weight-storage overhead. Future work could reduce this overhead through a sliding-window scheme that dynamically loads and evicts proxy weights across Transformer layers.

\section{Conclusion}

We present Prox, a training-free FFN sparsification framework for SwiGLU-based LLMs.
Prox employs a low-cost quantized proxy for input-dependent channel selection, followed by exact value computation of the selected channels.
Combined with a target-aware stage-wise allocation policy and tailored sparse kernels, Prox preserves model accuracy and achieves up to a $1.99\times$ end-to-end decoding speedup at 70\% FFN sparsity.
Comprehensive evaluations across ten LLMs from six model families demonstrate that Prox consistently outperforms existing training-free baselines across all sparsity levels and maintains seamless compatibility with model quantization and sparse attention.



\bibliography{Prox}

@article{arc,
  author  = {Clark, Peter and Cowhey, Isaac and Etzioni, Oren and Khot, Tushar and Sabharwal, Ashish and Schoenick, Carissa and Tafjord, Oyvind},
  journal = {arXiv preprint arXiv:1803.05457},
  title   = {Think you have solved question answering? try arc, the ai2 reasoning challenge},
  year    = {2018}
}

@article{awq,
  author  = {Lin, Ji and Tang, Jiaming and Tang, Haotian and Yang, Shang and Chen, Wei-Ming and Wang, Wei-Chen and Xiao, Guangxuan and Dang, Xingyu and Gan, Chuang and Han, Song},
  journal = {Proceedings of machine learning and systems},
  pages   = {87--100},
  title   = {Awq: Activation-aware weight quantization for on-device llm compression and acceleration},
  volume  = {6},
  year    = {2024}
}

@inproceedings{bisk2020piqa,
  author    = {Bisk, Yonatan and Zellers, Rowan and Gao, Jianfeng and Choi, Yejin and others},
  booktitle = {Proceedings of the AAAI conference on artificial intelligence},
  number    = {05},
  pages     = {7432--7439},
  title     = {Piqa: Reasoning about physical commonsense in natural language},
  volume    = {34},
  year      = {2020}
}

@article{cats,
  author  = {Lee, Donghyun and Lee, Je-Yong and Zhang, Genghan and Tiwari, Mo and Mirhoseini, Azalia},
  journal = {arXiv preprint arXiv:2404.08763},
  title   = {Cats: Contextually-aware thresholding for sparsity in large language models},
  year    = {2024}
}

@article{cobbe2021training,
  author  = {Cobbe, Karl and Kosaraju, Vineet and Bavarian, Mohammad and Chen, Mark and Jun, Heewoo and Kaiser, Lukasz and Plappert, Matthias and Tworek, Jerry and Hilton, Jacob and Nakano, Reiichiro and others},
  journal = {arXiv preprint arXiv:2110.14168},
  title   = {Training verifiers to solve math word problems},
  year    = {2021}
}

@inproceedings{countdown,
  author    = {Cheon, Jaewon and Kang, Pilsung},
  booktitle = {Proceedings of the 2025 Conference on Empirical Methods in Natural Language Processing},
  pages     = {28381--28397},
  title     = {COUNTDOWN: Contextually Sparse Activation Filtering Out Unnecessary Weights in Down Projection},
  year      = {2025}
}

@inproceedings{dauphin2017language,
  author       = {Dauphin, Yann N and Fan, Angela and Auli, Michael and Grangier, David},
  booktitle    = {International conference on machine learning},
  organization = {PMLR},
  pages        = {933--941},
  title        = {Language modeling with gated convolutional networks},
  year         = {2017}
}

@inproceedings{dejavu,
  author       = {Liu, Zichang and Wang, Jue and Dao, Tri and Zhou, Tianyi and Yuan, Binhang and Song, Zhao and Shrivastava, Anshumali and Zhang, Ce and Tian, Yuandong and Re, Christopher and others},
  booktitle    = {International Conference on Machine Learning},
  organization = {PMLR},
  pages        = {22137--22176},
  title        = {Deja vu: Contextual sparsity for efficient llms at inference time},
  year         = {2023}
}

@misc{eval-harness,
  author    = {Gao, Leo and Tow, Jonathan and Abbasi, Baber and Biderman, Stella and Black, Sid and DiPofi, Anthony and Foster, Charles and Golding, Laurence and Hsu, Jeffrey and Le Noac'h, Alain and Li, Haonan and McDonell, Kyle and Muennighoff, Niklas and Ociepa, Chris and Phang, Jason and Reynolds, Laria and Schoelkopf, Hailey and Skowron, Aviya and Sutawika, Lintang and Tang, Eric and Thite, Anish and Wang, Ben and Wang, Kevin and Zou, Andy},
  doi       = {10.5281/zenodo.12608602},
  month     = 07,
  publisher = {Zenodo},
  title     = {The Language Model Evaluation Harness},
  url       = {https://zenodo.org/records/12608602},
  version   = {v0.4.3},
  year      = 2024
}

@article{gemma_2025,
  author    = {Gemma Team},
  publisher = {Kaggle},
  title     = {Gemma 3},
  url       = {https://goo.gle/Gemma3Report},
  year      = {2025}
}

@article{gpt3_int8,
  author  = {Dettmers, Tim and Lewis, Mike and Belkada, Younes and Zettlemoyer, Luke},
  journal = {Advances in neural information processing systems},
  pages   = {30318--30332},
  title   = {Gpt3. int8 (): 8-bit matrix multiplication for transformers at scale},
  volume  = {35},
  year    = {2022}
}

@article{gptq,
  author  = {Frantar, Elias and Ashkboos, Saleh and Hoefler, Torsten and Alistarh, Dan},
  journal = {arXiv preprint arXiv:2210.17323},
  title   = {Gptq: Accurate post-training quantization for generative pre-trained transformers},
  year    = {2022}
}

@inproceedings{hshare,
  author    = {Wu, Huaijin and Li, Lianqiang and Huang, Hantao and Tu, Yi and Zhang, Jihang and Yu, Minghui and Yan, Junchi},
  booktitle = {The Thirteenth International Conference on Learning Representations},
  title     = {HShare: Fast LLM Decoding by Hierarchical Key-Value Sharing},
  year      = {2025}
}

@inproceedings{ig_pruning,
  author    = {Qiao, Kangyu and Zhang, Shaolei and Feng, Yang},
  booktitle = {Proceedings of the 2025 Conference on Empirical Methods in Natural Language Processing},
  pages     = {10629--10640},
  title     = {IG-Pruning: Input-Guided Block Pruning for Large Language Models},
  year      = {2025}
}

@misc{jiang2023mistral7b,
  archiveprefix = {arXiv},
  author        = {Albert Q. Jiang and Alexandre Sablayrolles and Arthur Mensch and Chris Bamford and Devendra Singh Chaplot and Diego de las Casas and Florian Bressand and Gianna Lengyel and Guillaume Lample and Lucile Saulnier and Lélio Renard Lavaud and Marie-Anne Lachaux and Pierre Stock and Teven Le Scao and Thibaut Lavril and Thomas Wang and Timothée Lacroix and William El Sayed},
  eprint        = {2310.06825},
  primaryclass  = {cs.CL},
  title         = {Mistral 7B},
  url           = {https://arxiv.org/abs/2310.06825},
  year          = {2023}
}

@inproceedings{kurtz2020inducing,
  author       = {Kurtz, Mark and Kopinsky, Justin and Gelashvili, Rati and Matveev, Alexander and Carr, John and Goin, Michael and Leiserson, William and Moore, Sage and Shavit, Nir and Alistarh, Dan},
  booktitle    = {International conference on machine learning},
  organization = {PMLR},
  pages        = {5533--5543},
  title        = {Inducing and exploiting activation sparsity for fast inference on deep neural networks},
  year         = {2020}
}

@article{lazy_neuron,
  author  = {Li, Zonglin and You, Chong and Bhojanapalli, Srinadh and Li, Daliang and Rawat, Ankit Singh and Reddi, Sashank J and Ye, Ke and Chern, Felix and Yu, Felix and Guo, Ruiqi and others},
  journal = {arXiv preprint arXiv:2210.06313},
  title   = {The lazy neuron phenomenon: On emergence of activation sparsity in transformers},
  year    = {2022}
}

@article{llama3modelcard,
  author = {AI@Meta},
  title  = {Llama 3 Model Card},
  url    = {https://github.com/meta-llama/llama3/blob/main/MODEL_CARD.md},
  year   = {2024}
}

@inproceedings{llm_in_a_flash,
  author    = {Alizadeh, Keivan and Mirzadeh, Seyed Iman and Belenko, Dmitry and Khatamifard, S and Cho, Minsik and Del Mundo, Carlo C and Rastegari, Mohammad and Farajtabar, Mehrdad},
  booktitle = {Proceedings of the 62nd Annual Meeting of the Association for Computational Linguistics (Volume 1: Long Papers)},
  pages     = {12562--12584},
  title     = {Llm in a flash: Efficient large language model inference with limited memory},
  year      = {2024}
}

@article{llm_pruner,
  author  = {Ma, Xinyin and Fang, Gongfan and Wang, Xinchao},
  journal = {Advances in neural information processing systems},
  pages   = {21702--21720},
  title   = {Llm-pruner: On the structural pruning of large language models},
  volume  = {36},
  year    = {2023}
}

@article{mmlu,
  author  = {Hendrycks, Dan and Burns, Collin and Basart, Steven and Zou, Andy and Mazeika, Mantas and Song, Dawn and Steinhardt, Jacob},
  journal = {arXiv preprint arXiv:2009.03300},
  title   = {Measuring massive multitask language understanding},
  year    = {2020}
}

@inproceedings{prosparse,
  author    = {Song, Chenyang and Han, Xu and Zhang, Zhengyan and Hu, Shengding and Shi, Xiyu and Li, Kuai and Chen, Chen and Liu, Zhiyuan and Li, Guangli and Yang, Tao and others},
  booktitle = {Proceedings of the 31st International Conference on Computational Linguistics},
  pages     = {2626--2644},
  title     = {Prosparse: Introducing and enhancing intrinsic activation sparsity within large language models},
  year      = {2025}
}

@misc{qwen3.5,
  author = {{Qwen Team}},
  month  = {February},
  title  = {{Qwen3.5}: Towards Native Multimodal Agents},
  url    = {https://qwen.ai/blog?id=qwen3.5},
  year   = {2026}
}

@misc{qwen3technicalreport,
  archiveprefix = {arXiv},
  author        = {Qwen Team},
  eprint        = {2505.09388},
  primaryclass  = {cs.CL},
  title         = {Qwen3 Technical Report},
  url           = {https://arxiv.org/abs/2505.09388},
  year          = {2025}
}

@article{raihan2020sparse,
  author  = {Raihan, Md Aamir and Aamodt, Tor},
  journal = {Advances in Neural Information Processing Systems},
  pages   = {15625--15638},
  title   = {Sparse weight activation training},
  volume  = {33},
  year    = {2020}
}

@inproceedings{relu_strikes_back,
  author    = {Mirzadeh, Iman and Alizadeh-Vahid, Keivan and Mehta, Sachin and Del Mundo, Carlo C and Tuzel, Oncel and Samei, Golnoosh and Rastegari, Mohammad and Farajtabar, Mehrdad},
  booktitle = {International Conference on Learning Representations},
  pages     = {25378--25400},
  title     = {Relu strikes back: Exploiting activation sparsity in large language models},
  volume    = {2024},
  year      = {2024}
}

@article{relu2_wins,
  author  = {Zhang, Zhengyan and Song, Yixin and Yu, Guanghui and Han, Xu and Lin, Yankai and Xiao, Chaojun and Song, Chenyang and Liu, Zhiyuan and Mi, Zeyu and Sun, Maosong},
  journal = {arXiv preprint arXiv:2402.03804},
  title   = {ReLU$^2$ Wins: Discovering Efficient Activation Functions for Sparse LLMs},
  year    = {2024}
}

@inproceedings{rocketkv,
  author    = {Payman Behnam and Yaosheng Fu and Ritchie Zhao and Po-An Tsai and Zhiding Yu and Alexey Tumanov},
  booktitle = {Forty-second International Conference on Machine Learning},
  title     = {Rocket{KV}: Accelerating Long-Context {LLM} Inference via Two-Stage {KV} Cache Compression},
  url       = {https://openreview.net/forum?id=RyOpooIxDF},
  year      = {2025}
}

@article{shazeer2020glu,
  author  = {Shazeer, Noam},
  journal = {arXiv preprint arXiv:2002.05202},
  title   = {GLU Variants Improve Transformer},
  year    = {2020}
}

@inproceedings{shortgpt,
  author    = {Men, Xin and Xu, Mingyu and Zhang, Qingyu and Yuan, Qianhao and Wang, Bingning and Lin, Hongyu and Lu, Yaojie and Han, Xianpei and Chen, Weipeng},
  booktitle = {Findings of the Association for Computational Linguistics: ACL 2025},
  pages     = {20192--20204},
  title     = {Shortgpt: Layers in large language models are more redundant than you expect},
  year      = {2025}
}

@inproceedings{slicegpt,
  author    = {Ashkboos, Saleh and Croci, Maximilian and Gennari do Nascimento, Marcelo and Hoefler, Torsten and Hensman, James},
  booktitle = {International Conference on Learning Representations},
  pages     = {11682--11701},
  title     = {Slicegpt: Compress large language models by deleting rows and columns},
  volume    = {2024},
  year      = {2024}
}

@inproceedings{sparsegpt,
  author       = {Frantar, Elias and Alistarh, Dan},
  booktitle    = {International conference on machine learning},
  organization = {PMLR},
  pages        = {10323--10337},
  title        = {Sparsegpt: Massive language models can be accurately pruned in one-shot},
  year         = {2023}
}

@inproceedings{sparsevit,
  author    = {Chen, Xuanyao and Liu, Zhijian and Tang, Haotian and Yi, Li and Zhao, Hang and Han, Song},
  booktitle = {Proceedings of the IEEE/CVF conference on computer vision and pattern recognition},
  pages     = {2061--2070},
  title     = {Sparsevit: Revisiting activation sparsity for efficient high-resolution vision transformer},
  year      = {2023}
}

@inproceedings{sun2024simple,
  author    = {Sun, Mingjie and Liu, Zhuang and Bair, Anna and Kolter, Zico},
  booktitle = {International Conference on Learning Representations},
  pages     = {4942--4964},
  title     = {A simple and effective pruning approach for large language models},
  volume    = {2024},
  year      = {2024}
}

@inproceedings{teal,
  author    = {Liu, James and Ponnusamy, Pragaash and Cai, Tianle and Kim, Yoon and Athiwaratkun, Ben and others},
  booktitle = {International Conference on Learning Representations},
  pages     = {98302--98322},
  title     = {Training-free activation sparsity in large language models},
  volume    = {2025},
  year      = {2025}
}

@inproceedings{triton,
  author    = {Tillet, Philippe and Kung, Hsiang-Tsung and Cox, David},
  booktitle = {Proceedings of the 3rd ACM SIGPLAN International Workshop on Machine Learning and Programming Languages},
  pages     = {10--19},
  title     = {Triton: an intermediate language and compiler for tiled neural network computations},
  year      = {2019}
}

@article{turbo_sparse,
  author  = {Song, Yixin and Xie, Haotong and Zhang, Zhengyan and Wen, Bo and Ma, Li and Mi, Zeyu and Chen, Haibo},
  journal = {arXiv preprint arXiv:2406.05955},
  title   = {Turbo sparse: Achieving llm sota performance with minimal activated parameters},
  year    = {2024}
}

@misc{universal_properties_activation_sparsity,
  archiveprefix = {arXiv},
  author        = {Filip Szatkowski and Patryk Będkowski and Alessio Devoto and Jan Dubiński and Pasquale Minervini and Mikołaj Piórczyński and Simone Scardapane and Bartosz Wójcik},
  eprint        = {2509.00454},
  primaryclass  = {cs.LG},
  title         = {Universal Properties of Activation Sparsity in Modern Large Language Models},
  url           = {https://arxiv.org/abs/2509.00454},
  year          = {2026}
}

@article{wikitext,
  author  = {Merity, Stephen and Xiong, Caiming and Bradbury, James and Socher, Richard},
  journal = {arXiv preprint arXiv:1609.07843},
  title   = {Pointer sentinel mixture models},
  year    = {2016}
}

@inproceedings{zellers2019hellaswag,
  author    = {Zellers, Rowan and Holtzman, Ari and Bisk, Yonatan and Farhadi, Ali and Choi, Yejin},
  booktitle = {Proceedings of the 57th annual meeting of the association for computational linguistics},
  pages     = {4791--4800},
  title     = {Hellaswag: Can a machine really finish your sentence?},
  year      = {2019}
}

@article{zhang2025rsparse,
  author  = {Zhang, Zhenyu and Liu, Zechun and Tian, Yuandong and Khaitan, Harshit and Wang, Zhangyang and Li, Steven},
  journal = {arXiv preprint arXiv:2504.19449},
  title   = {R-sparse: Rank-aware activation sparsity for efficient llm inference},
  year    = {2025}
}

\clearpage
\appendix
\setcounter{secnumdepth}{2}
\section{Appendix}

\subsection{FFN Parameter and MAC Calculation}
\label{app:ffn_calculation}

We calculate the parameter and MAC shares in Figure~\ref{fig:qwen3_ffn_block_shares} as follows.
Let $S$ denote the sequence length, $d_{\mathrm{model}}$ the hidden dimension, $d_{\mathrm{ff}}$ the FFN intermediate dimension, $H_q$ and $H_{\mathrm{kv}}$ the numbers of query and key--value heads, and $d_h$ the head dimension.
The query and key--value projection dimensions are defined as:
\[
  d_q=H_qd_h,
  \qquad
  d_{\mathrm{kv}}=H_{\mathrm{kv}}d_h.
\]
The parameter counts of the FFN and one Transformer block are
\[
  P_{\mathrm{FFN}}=3d_{\mathrm{model}}d_{\mathrm{ff}},
\]
\[
  \begin{array}{rcl}
    P_{\mathrm{Block}}
    &=&3d_{\mathrm{model}}d_{\mathrm{ff}}
    +2d_{\mathrm{model}}(d_q+d_{\mathrm{kv}})\\
    &&{}+2d_{\mathrm{model}}+2d_h,
  \end{array}
\]
\[
  R_{\mathrm{param}}=\frac{P_{\mathrm{FFN}}}{P_{\mathrm{Block}}}.
\]
At sequence length $S$, the corresponding MAC counts are
\[
  M_{\mathrm{FFN}}=3Sd_{\mathrm{model}}d_{\mathrm{ff}},
\]
\[
  \begin{array}{rcl}
    M_{\mathrm{Block}}
    &=&3Sd_{\mathrm{model}}d_{\mathrm{ff}}
    +2Sd_{\mathrm{model}}(d_q+d_{\mathrm{kv}})\\
    &&{}+2S^2d_q,
  \end{array}
\]
\[
  R_{\mathrm{MAC}}=\frac{M_{\mathrm{FFN}}}{M_{\mathrm{Block}}}.
\]
We set $S=4096$ and use the model configurations in Table~\ref{tab:qwen3_block_configs}.
The calculation includes the Q/K/V/O projections and RMSNorm parameters, but excludes embeddings, the LM head, softmax, RoPE, activation functions, and other element-wise operations.

\subsection{Activation Distributions and Ranking Preservation}
\label{app:activation_distributions}

\begin{figure*}[t]
  \centering
  \includegraphics[width=\textwidth]{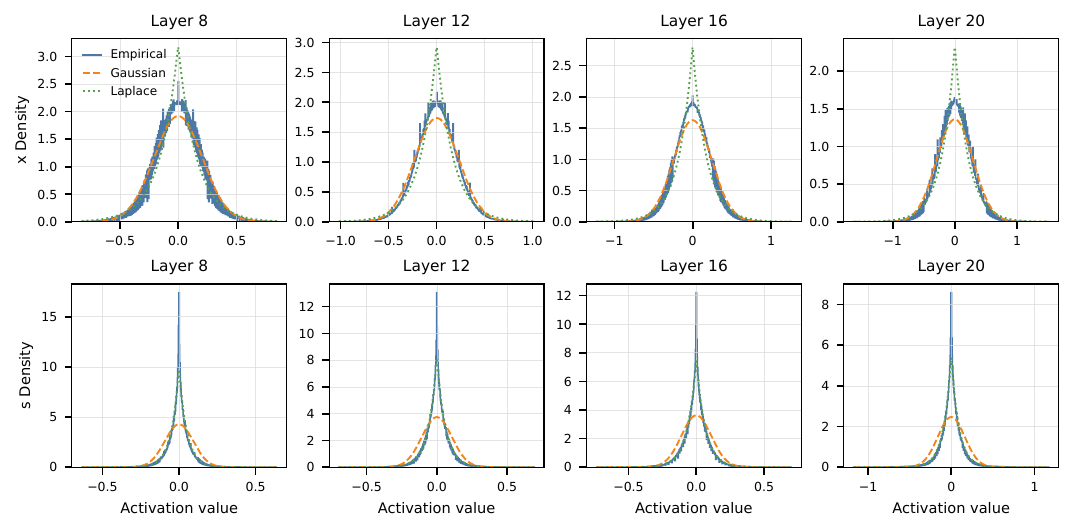}
  \caption{Activation-value distributions at representative SwiGLU FFN layers of Qwen3-8B, collected using the same calibration set as threshold calibration: the first 10 contiguous sequences of 2,048 tokens from the Alpaca training split. The top row shows the shared input $\mathbf{x}$ to the up and gate projections, together with fitted Gaussian and Laplace densities. The bottom row shows the intermediate state $\mathbf{s}$ supplied to the down projection. The FFN input is Gaussian-like, whereas the intermediate state is Laplacian-like and more sharply concentrated around zero.}
  \label{fig:activation_value_distributions}
\end{figure*}

Figure~\ref{fig:activation_value_distributions} contrasts the activation distributions at the two sides of a SwiGLU FFN. The FFN input $\mathbf{x}$, shared by the up and gate projections, is approximately Gaussian, whereas the intermediate state $\mathbf{s}$ supplied to the down projection is approximately Laplacian. These distinct shapes clarify why magnitude sparsification plays different roles in exact-$\mathbf{s}$ channel selection, TEAL's successive input-sparsity scheme, and the selection-oriented proxy used by Prox.

\textbf{Intermediate-state compressibility.}
The Laplacian-like shape of $\mathbf{s}$ consists of a sharp central peak and comparatively heavy tails, so many channels have negligible magnitudes while a small fraction carries most of the activation energy. This concentration can be quantified under an ideal Laplace model. Let $S\sim\mathrm{Laplace}(0,b)$, and prune the fraction $p$ of entries with the smallest magnitudes. With $a_p=-\log(1-p)$, the retained fraction of squared activation energy is
\[
  q_{\mathrm{Lap}}(p)
  =
  \frac{\mathrm{E}\!\left[S^2\mathbf{1}(|S|>ba_p)\right]}
  {\mathrm{E}[S^2]}
  =
  \frac{1-p}{2}\left(a_p^2+2a_p+2\right).
\]
At $p=0.7$, retaining only the largest-magnitude 30\% of entries preserves approximately 87.9\% of the squared activation energy. Although activation energy alone does not determine the final FFN error because each coefficient $s_i$ scales a different row of $W_{\mathrm{down}}$, this calculation provides a distributional explanation consistent with the oracle-$\mathbf{s}$ result in Figure~\ref{fig:oracle_s_ppl_vs_sparsity}(a): most low-magnitude intermediate channels can be removed with little quality degradation.

\begin{table}[t]
  \centering
  \footnotesize
  \setlength{\tabcolsep}{2.4pt}
  \renewcommand{\arraystretch}{1.08}
  \begin{tabular*}{\columnwidth}{@{\extracolsep{\fill}}lrrrrr@{}}
    \toprule
    \textbf{Model} & $d_{\mathrm{model}}$ & $d_{\mathrm{ff}}$ & $H_q$ & $H_{\mathrm{kv}}$ & $d_h$ \\
    \midrule
    Qwen3-1.7B & 2048 & 6144  & 16 & 8 & 128 \\
    Qwen3-4B   & 2560 & 9728  & 32 & 8 & 128 \\
    Qwen3-8B   & 4096 & 12288 & 32 & 8 & 128 \\
    Qwen3-14B  & 5120 & 17408 & 40 & 8 & 128 \\
    Qwen3-32B  & 5120 & 25600 & 64 & 8 & 128 \\
    \bottomrule
  \end{tabular*}
  \caption{Qwen3 configurations used to calculate the FFN parameter and MAC shares in Figure~\ref{fig:qwen3_ffn_block_shares}.}
  \label{tab:qwen3_block_configs}
\end{table}

\textbf{Value distortion under input sparsity.}
The Gaussian-like FFN input is less concentrated than the Laplacian-like intermediate state, so magnitude pruning at the same sparsity discards a larger fraction of its energy. Following the idealized analysis of magnitude-based activation sparsity in TEAL~\citep{teal}, let $X\sim\mathcal{N}(0,\sigma_x^2)$, assume independent isotropic Gaussian weights, and prune the fraction $p$ of input entries satisfying $|X|\leq \sigma_x z_p$, where $z_p=\Phi^{-1}((1+p)/2)$, and $\Phi$ and $\phi$ denote the standard Gaussian cumulative distribution and density functions. The retained input-energy fraction is
\[
  q_{\mathrm{G}}(p)
  =1-p+2z_p\phi(z_p),
\]
and the corresponding relative root-mean-square projection error is
\[
  \eta(p)=\sqrt{1-q_{\mathrm{G}}(p)}
  =\sqrt{p-2z_p\phi(z_p)}.
\]
For example, at 70\% input sparsity, $q_{\mathrm{G}}(0.7)\approx0.783$ and $\eta(0.7)\approx0.465$. Thus, magnitude sparsification is substantially better than discarding inputs indiscriminately, but its projected values can still differ considerably from their dense counterparts.
In methods such as TEAL with successive input-sparsity stages, approximation errors accumulate as they propagate through the up and gate branches, their element-wise product, and the down projection, making aggressive input sparsity increasingly fragile.

\textbf{Ranking preservation under input sparsity.}
Large value error does not imply that the proxy has lost its ranking information. For one linear-projection output channel with weight vector $\mathbf{w}_j$, decompose $\mathbf{x}=\mathbf{x}_{\mathrm{sp}}+\mathbf{r}$ into the retained input and discarded residual:
\[
  y_j
  =\langle\mathbf{x},\mathbf{w}_j\rangle
  =\underbrace{\langle\mathbf{x}_{\mathrm{sp}},\mathbf{w}_j\rangle}_{\tilde{y}_j}
  +\underbrace{\langle\mathbf{r},\mathbf{w}_j\rangle}_{e_j}.
\]
Under the same independence and isotropy assumptions, $\tilde{y}_j$ and $e_j$ are uncorrelated. Consequently,
\[
  \mathrm{Corr}(y_j,\tilde{y}_j)
  =
  \sqrt{\frac{\mathrm{Var}(\tilde{y}_j)}
  {\mathrm{Var}(y_j)}}
  \approx\sqrt{q_{\mathrm{G}}(p)}.
\]
At 70\% input sparsity, this idealized correlation is approximately $0.885$, despite the relative projection error of 0.465. Applying the same sparse input to both SwiGLU branches therefore preserves a substantial shared signal from which Prox constructs $\tilde{\mathbf{s}}$. If $\tilde{s}_i=s_i+\delta_i$, the magnitude order of channels $i$ and $j$ is guaranteed to remain unchanged whenever
\[
  \bigl||s_i|-|s_j|\bigr|>|\delta_i|+|\delta_j|.
\]
The large-magnitude tail channels of the Laplacian-like $\mathbf{s}$ can remain well separated from the selection boundary, while ranking disagreements concentrate among channels near that boundary. The Gaussian calculation does not model SiLU, the product of the two branches, trained-weight dependencies, or proxy-weight quantization, so it should be interpreted as a mechanism rather than a guarantee. The observed 82.04\% exact--proxy selection overlap in Figure~\ref{fig:oracle_s_ppl_vs_sparsity}(d) verifies that this mechanism persists in the complete proxy. Prox exploits this weaker ranking requirement and computes the retained values exactly, whereas TEAL propagates the approximate values themselves.

\subsection{Empirical Proxy-Cost Ratios}
\label{app:alpha_validation}

The allocation model uses $\alpha$ to approximate the cost of one INT4 proxy GEMV relative to an FP16 GEMV at the same input sparsity. We measure this latency ratio across NVIDIA A6000, GeForce RTX 3090, and A100 GPUs for four Qwen3 scales and four input sparsity levels, as reported in Table~\ref{tab:alpha_measurements}.

The measured ratios range from 0.234 to 0.351, with a pooled mean of 0.285. We use the single rounded value $\alpha=1/3$, which falls within the observed range and avoids device- or model-specific allocation tuning.

\begin{table}[t]
  \centering
  \small
  \setlength{\tabcolsep}{3pt}
  \renewcommand{\arraystretch}{1.08}
  \begin{tabular*}{\columnwidth}{@{\extracolsep{\fill}}lcccc@{}}
    \toprule
    \textbf{Model} & \multicolumn{4}{c}{\textbf{Input sparsity}} \\
    \cmidrule(l){2-5}
    & \textbf{0\%} & \textbf{10\%} & \textbf{55\%} & \textbf{70\%} \\
    \midrule
    \multicolumn{5}{l}{\textbf{NVIDIA A6000}} \\
    Qwen3-4B  & 0.280 & 0.292 & 0.304 & 0.286 \\
    Qwen3-8B  & 0.267 & 0.272 & 0.286 & 0.277 \\
    Qwen3-14B & 0.261 & 0.273 & 0.287 & 0.292 \\
    Qwen3-32B & 0.255 & 0.263 & 0.268 & 0.262 \\
    \addlinespace[0.15em]
    \multicolumn{5}{l}{\textbf{NVIDIA GeForce RTX 3090}} \\
    Qwen3-4B  & 0.300 & 0.311 & 0.328 & 0.309 \\
    Qwen3-8B  & 0.279 & 0.298 & 0.310 & 0.292 \\
    Qwen3-14B & 0.274 & 0.291 & 0.316 & 0.323 \\
    Qwen3-32B & 0.261 & 0.278 & 0.290 & 0.281 \\
    \addlinespace[0.15em]
    \multicolumn{5}{l}{\textbf{NVIDIA A100}} \\
    Qwen3-4B  & 0.234 & 0.292 & 0.274 & 0.273 \\
    Qwen3-8B  & 0.281 & 0.341 & 0.261 & 0.248 \\
    Qwen3-14B & 0.279 & 0.351 & 0.320 & 0.241 \\
    Qwen3-32B & 0.272 & 0.313 & 0.288 & 0.239 \\
    \bottomrule
  \end{tabular*}
  \caption{Measured proxy-cost ratio $\alpha$: latency of the INT4 proxy GEMV divided by that of the FP16 GEMV at the same input sparsity. Values are reported across GPUs, Qwen3 scales, and input sparsities.}
  \label{tab:alpha_measurements}
\end{table}

\subsection{Stage-Wise Compute Configurations}
\label{app:stagewise_configs}

Table~\ref{tab:stagewise_allocation} lists the Stage~1 and Stage~2 sparsities used to realize each target effective sparsity $e$ in our experiments.
Applying the allocation rule with $s_{\mathrm{ref}}=0.7$ and $\alpha=1/3$ keeps $s_2$ at 70\% for the 50\% and 60\% operating points, with $s_1$ absorbing the remaining budget.
At the 40\% and 70\% operating points, the Stage~1 allocation implied by $s_2=s_{\mathrm{ref}}$ falls outside the allowed range, so $s_1$ is clamped to 0\% and 70\%, respectively; $s_2$ is then adjusted to 62.2\% and 76.7\% so that each configuration still satisfies its target $e$.

\begin{table}[t]
  \centering
  \small
  \setlength{\tabcolsep}{4pt}
  \renewcommand{\arraystretch}{1.08}
  \begin{tabular*}{\columnwidth}{@{\extracolsep{\fill}}ccc@{}}
    \toprule
    \textbf{Target $e$}
    & \textbf{\shortstack{Stage~1\\sparsity $s_1$}}
    & \textbf{\shortstack{Stage~2\\sparsity $s_2$}} \\
    \midrule
    40\% & 0\%  & 62.2\% \\
    50\% & 10\% & 70.0\% \\
    60\% & 55\% & 70.0\% \\
    70\% & 70\% & 76.7\% \\
    \bottomrule
  \end{tabular*}
  \caption{Stage-wise allocations used for each target sparsity. The configurations follow the proposed allocation rule with $s_{\mathrm{ref}}=0.7$, $s_1\in[0,0.7]$, and $\alpha=1/3$.}
  \label{tab:stagewise_allocation}
\end{table}

\subsection{End-to-End Latency Analysis}
\label{app:e2e_3090}

\begin{figure}[t]
  \centering
  \includegraphics[width=\columnwidth]{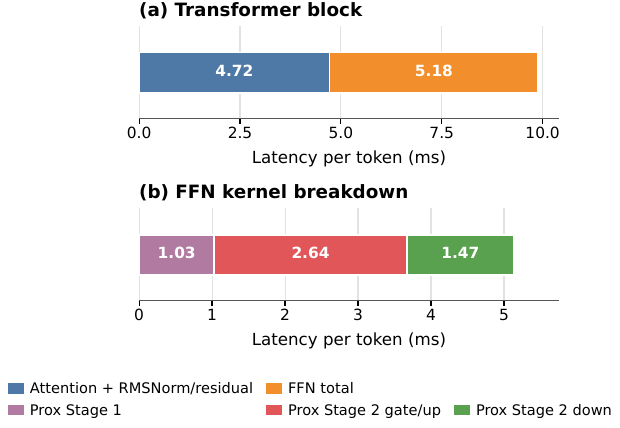}
  \caption{Detailed latency breakdown of Prox on Qwen3-8B with 70\% effective FFN sparsity on an NVIDIA A6000 GPU. Values are normalized per generated token over 36 Transformer layers. (a) The Transformer-block profile totals 9.88~ms/token. (b) The three FFN kernels are independently profiled, causing a 0.04~ms/token difference from the aggregate FFN timing. Attention, RMSNorm, and residual are reported jointly because Inductor fuses them into one kernel.}
  \label{fig:latency_breakdown}
\end{figure}

Figure~\ref{fig:latency_breakdown} provides a detailed latency breakdown for Prox on Qwen3-8B at 70\% sparsity. Figure~\ref{fig:latency_breakdown}(a) attributes 4.72~ms/token to the fused attention, RMSNorm, and residual path and 5.18~ms/token to the FFN, versus a 9.88~ms/token Transformer-block total. The small aggregation difference arises because the regions are profiled independently.

Figure~\ref{fig:latency_breakdown}(b) further decomposes the FFN: Stage~2 gate/up is the largest component at 2.64~ms/token (51.0\% of FFN time), while Stage~1 and Stage~2 down take 1.03 and 1.47~ms/token, respectively. The three kernel timings sum to 5.13~ms/token, 0.04~ms/token (0.8\%) below the aggregate FFN timing because they are independently profiled.

We evaluate end-to-end decoding throughput on NVIDIA A100, A6000, and GeForce RTX 3090 GPUs using six representative models, with unmeasured model--device configurations marked by dashes.
Table~\ref{tab:e2e_speed_3090} reports the dense baseline and the resulting throughput for Prox and TEAL at each target FFN sparsity.

\begin{table*}[t]
  \centering
  \small
  \setlength{\tabcolsep}{1mm}
  \renewcommand{\arraystretch}{1.08}
  \begin{tabular*}{\textwidth}{@{\extracolsep{\fill}}ll*{15}{c}@{}}
    \toprule
    & & \multicolumn{5}{c}{\textbf{NVIDIA A100}} & \multicolumn{5}{c}{\textbf{NVIDIA A6000}} & \multicolumn{5}{c}{\textbf{NVIDIA GeForce RTX 3090}} \\
    \cmidrule(lr){3-7}\cmidrule(lr){8-12}\cmidrule(lr){13-17}
    \textbf{Model} & \textbf{Method}
    & \textbf{Dense} & \textbf{40\%} & \textbf{50\%} & \textbf{60\%} & \textbf{70\%}
    & \textbf{Dense} & \textbf{40\%} & \textbf{50\%} & \textbf{60\%} & \textbf{70\%}
    & \textbf{Dense} & \textbf{40\%} & \textbf{50\%} & \textbf{60\%} & \textbf{70\%} \\
    \midrule
    \multirow{2}{*}{Qwen3-4B} & Prox
    & \multirow{2}{*}{152.1} & 169.7 & 169.9 & 179.6 & 184.8
    & \multirow{2}{*}{82.0} & 107.7 & 116.0 & 123.6 & 135.2
    & \multirow{2}{*}{95.5} & 121.8 & 130.8 & 140.4 & 150.1 \\
    & TEAL
    & & 167.2 & 175.9 & 185.8 & 194.3 & & 105.4 & 113.3 & 123.2 & 134.7 & & 119.1 & 125.8 & 137.1 & 149.1 \\
    \addlinespace[0.15em]
    \multirow{2}{*}{Qwen3-8B} & Prox
    & \multirow{2}{*}{93.5} & 113.0 & 117.3 & 125.3 & 133.5
    & \multirow{2}{*}{46.7} & 65.2 & 70.9 & 78.0 & 85.0
    & \multirow{2}{*}{54.7} & 74.1 & 80.6 & 86.3 & 95.1 \\
    & TEAL
    & & 110.6 & 118.8 & 127.6 & 135.2 & & 63.1 & 69.3 & 76.2 & 86.8 & & 73.5 & 77.4 & 86.1 & 95.7 \\
    \addlinespace[0.15em]
    \multirow{2}{*}{Qwen3-14B} & Prox
    & \multirow{2}{*}{53.1} & 69.4 & 71.4 & 77.5 & 82.4
    & \multirow{2}{*}{25.8} & 37.0 & 41.5 & 44.5 & 50.9
    & \multirow{2}{*}{--} & -- & -- & -- & -- \\
    & TEAL
    & & 65.8 & 71.0 & 77.8 & 83.1 & & 35.5 & 39.6 & 43.9 & 51.1 & & -- & -- & -- & -- \\
    \addlinespace[0.15em]
    \multirow{2}{*}{Qwen3-32B} & Prox
    & \multirow{2}{*}{23.9} & 32.0 & 33.4 & 36.7 & 39.3
    & \multirow{2}{*}{--} & -- & -- & -- & --
    & \multirow{2}{*}{--} & -- & -- & -- & -- \\
    & TEAL
    & & 30.5 & 33.1 & 36.0 & 39.1 & & -- & -- & -- & -- & & -- & -- & -- & -- \\
    \addlinespace[0.15em]
    \multirow{2}{*}{Llama3-8B} & Prox
    & \multirow{2}{*}{94.4} & 120.4 & 122.6 & 130.3 & 140.7
    & \multirow{2}{*}{47.3} & 66.8 & 74.1 & 80.7 & 89.6
    & \multirow{2}{*}{55.3} & 77.0 & 84.1 & 90.6 & 101.5 \\
    & TEAL
    & & 114.8 & 122.7 & 132.7 & 142.3 & & 64.7 & 71.2 & 78.5 & 89.6 & & 72.7 & 78.1 & 87.2 & 100.0 \\
    \addlinespace[0.15em]
    \multirow{2}{*}{Mistral-7B} & Prox
    & \multirow{2}{*}{100.3} & 127.5 & 131.7 & 142.0 & 152.9
    & \multirow{2}{*}{50.3} & 73.1 & 80.3 & 88.1 & 100.3
    & \multirow{2}{*}{58.6} & 83.1 & 91.6 & 100.6 & 112.1 \\
    & TEAL
    & & 123.0 & 134.1 & 147.2 & 158.3 & & 70.0 & 78.7 & 88.2 & 103.2 & & 78.4 & 85.8 & 98.9 & 116.7 \\
    \bottomrule
  \end{tabular*}
  \caption{Steady-state single-batch end-to-end decoding throughput on NVIDIA A100, A6000, and GeForce RTX 3090 GPUs. Entries report tokens per second; dashes indicate unmeasured model--device configurations.}
  \label{tab:e2e_speed_3090}
\end{table*}

Across all three GPUs, Prox and TEAL deliver broadly comparable throughput over the evaluated models and sparsity levels, with neither method consistently dominating.

\subsection{Threshold-Based Selection and Token-Adaptive Computation}
\label{app:threshold_selection}

Prox converts two continuous magnitude signals into binary masks: the FFN input $\mathbf{x}$ determines the proxy-path input mask $\mathbf{m}_{\mathbf{x}}$, while the proxy intermediate state $\tilde{\mathbf{s}}$ determines the final intermediate-channel mask $\mathbf{m}_{\mathbf{s}}$.
We use layer-wise magnitude thresholds for both signals instead of selecting a fixed number of coordinates independently for each token.

\textbf{Per-token top-$k$ selection.}
Given a salience signal $\mathbf{z}$ for one token, top-$k$ selection retains the indices of its $k$ largest-magnitude entries:
\[
  [\mathbf{m}^{\mathrm{top}\mbox{-}k}(\mathbf{z})]_i
  =\mathbf{1}\!\left[
    i\in\mathrm{TopKIndices}(|\mathbf{z}|,k)
  \right].
\]
This rule strictly controls the number of retained coordinates for every token, but uses only the magnitude ranking within that token.
Consequently, two tokens receive the same compute budget even when their overall activation scales or importance distributions differ substantially.
Top-$k$ therefore enforces a per-token budget rather than a shared absolute importance criterion.

\textbf{Layer-wise threshold selection.}
For FFN layer $\ell$, a threshold rule instead retains every coordinate whose magnitude exceeds a fixed layer-specific value:
\[
  [\mathbf{m}^{\tau}(\mathbf{z})]_i
  =\mathbf{1}\!\left[|z_i|\geq\tau_{\ell}\right].
\]
Unlike top-$k$, this rule does not force all tokens to retain the same number of coordinates.
Tokens with fewer entries above the threshold use less computation, whereas tokens with more salient entries retain more coordinates.
Thresholding therefore controls the average compute budget while allowing that budget to vary with the token-specific activation distribution.
It also avoids a runtime per-token ranking operation.

\textbf{Measuring token-adaptive computation.}
For token $t$ at layer $\ell$, Prox applies this rule separately to the input and proxy intermediate signals:
\[
  [\mathbf{m}_{\mathbf{x},\ell,t}]_i
  =\mathbf{1}\!\left[
    |x_{\ell,t,i}|\geq\tau_{\mathbf{x},\ell}^{s_1}
  \right],
\]
\[
  [\mathbf{m}_{\mathbf{s},\ell,t}]_j
  =\mathbf{1}\!\left[
    |\tilde{s}_{\ell,t,j}|\geq\tau_{\mathbf{s},\ell}^{s_2}
  \right].
\]
The thresholds $\tau_{\mathbf{x},\ell}^{s_1}$ and $\tau_{\mathbf{s},\ell}^{s_2}$ are calibrated to meet the target Stage~1 and Stage~2 sparsities on average, rather than enforcing those sparsities independently for every token.
We define the token-wise sparsities induced by the two masks as
\[
  r_{\ell,t}^{\mathbf{x}}
  =1-\frac{\|\mathbf{m}_{\mathbf{x},\ell,t}\|_0}{d_{\mathrm{model}}},
  \qquad
  r_{\ell,t}^{\tilde{\mathbf{s}}}
  =1-\frac{\|\mathbf{m}_{\mathbf{s},\ell,t}\|_0}{d_{\mathrm{ff}}}.
\]
We measure their variation at two scales.
For the within-layer view, we compute the standard deviation and the 10th and 90th percentiles across tokens in each FFN layer, then average each statistic across layers.
For the across-layer view, we first average each token's sparsity over the $L$ FFN layers,
\[
  \bar r_t^{\mathbf{x}}
  =\frac{1}{L}\sum_{\ell=1}^{L}r_{\ell,t}^{\mathbf{x}},
  \qquad
  \bar r_t^{\tilde{\mathbf{s}}}
  =\frac{1}{L}\sum_{\ell=1}^{L}r_{\ell,t}^{\tilde{\mathbf{s}}},
\]
and then compute the corresponding statistics across tokens.

\begin{table}[t]
  \centering
  \footnotesize
  \setlength{\tabcolsep}{3.0pt}
  \renewcommand{\arraystretch}{1.08}
  \begin{tabular*}{\columnwidth}{@{\extracolsep{\fill}}llccc@{}}
    \toprule
    \textbf{Aggregation} & \textbf{Signal} & \textbf{Std. (pts)} & \textbf{P10} & \textbf{P90} \\
    \midrule
    \multirow{2}{*}{Within-layer}
    & $\mathbf{x}$ mask & 6.90 & 60.49\% & 77.82\% \\
    & $\tilde{\mathbf{s}}$ mask & 6.80 & 63.83\% & 81.34\% \\
    \addlinespace[0.15em]
    \multirow{2}{*}{Across-layer}
    & $\mathbf{x}$ mask & 4.78 & 63.16\% & 75.23\% \\
    & $\tilde{\mathbf{s}}$ mask & 4.60 & 67.05\% & 79.01\% \\
    \bottomrule
  \end{tabular*}
  \caption{Token-wise sparsity variation under layer-wise threshold selection. Within-layer statistics are computed across tokens per layer and then averaged across layers; across-layer statistics are computed after averaging each token's sparsity over all FFN layers.}
  \label{tab:token_adaptive_sparsity}
\end{table}

Table~\ref{tab:token_adaptive_sparsity} shows substantial variation in the compute allocated to different tokens.
Within individual layers, the P10--P90 ranges span 17.33 percentage points for the input mask and 17.51 points for the proxy-intermediate mask.
Even after averaging over all FFN layers, the corresponding ranges remain 12.07 and 11.96 points.
In contrast, a fixed per-token top-$k$ rule has zero token-wise sparsity variation by construction.
These results confirm that threshold selection gives Prox a natural token-adaptive computation mechanism instead of assigning every token the same retained-channel count.

\textbf{Thresholding versus top-$k$.}
Having verified that thresholding enables token-adaptive computation, we directly compare the two selection rules on Qwen3-8B and Qwen3-14B at matched target effective sparsities.
Table~\ref{tab:threshold_vs_topk} reports the aggregate downstream scores.

\begin{table}[t]
  \centering
  \footnotesize
  \setlength{\tabcolsep}{4.0pt}
  \renewcommand{\arraystretch}{1.08}
  \begin{tabular*}{\columnwidth}{@{\extracolsep{\fill}}lccc@{}}
    \toprule
    \textbf{Model} & \textbf{Eff. sparsity} & \textbf{Threshold} & \textbf{Top-$k$} \\
    \midrule
    \multirow{4}{*}{Qwen3-8B}
    & 40\% & \textbf{75.70} & 73.04 \\
    & 50\% & \textbf{74.60} & 71.62 \\
    & 60\% & \textbf{73.30} & 70.49 \\
    & 70\% & \textbf{68.60} & 67.57 \\
    \addlinespace[0.15em]
    \multirow{4}{*}{Qwen3-14B}
    & 40\% & \textbf{77.60} & 77.00 \\
    & 50\% & \textbf{76.60} & 74.72 \\
    & 60\% & \textbf{76.10} & 74.81 \\
    & 70\% & \textbf{74.80} & 73.74 \\
    \bottomrule
  \end{tabular*}
  \caption{Aggregate downstream scores using layer-wise threshold selection and per-token top-$k$ at matched target effective sparsity. Higher is better; bold marks the better selection rule in each setting.}
  \label{tab:threshold_vs_topk}
\end{table}

Thresholding achieves a higher score in all eight settings and improves the average score by 1.79 points over top-$k$.
On Qwen3-8B, its gains range from 1.03 to 2.98 points across the four sparsity levels.
On Qwen3-14B, thresholding is consistently better by 0.60--1.88 points.
Thus, preserving an absolute layer-wise importance criterion consistently improves downstream quality across the evaluated settings at a matched average compute budget.
Having established the motivation and empirical benefits of thresholding, Appendix~\ref{app:calibration_runtime_sparsity} describes how the layer-wise thresholds are calibrated and evaluates their runtime sparsity stability.

\subsection{Threshold Calibration and Runtime Sparsity}
\label{app:calibration_runtime_sparsity}

\textbf{Calibration data and scale.}
Prox calibrates its fixed layer-specific thresholds using generic instruction text from the Alpaca training split.
For each model, we tokenize the calibration corpus with its native tokenizer, partition the resulting tokens into contiguous sequences of length 2,048, and use the first 10 sequences for each target-sparsity configuration.
Thus, each configuration is calibrated on 20,480 token positions.
We independently calibrate the ten models in our main evaluation---Qwen3-1.7B, Qwen3-4B, Qwen3-8B, Qwen3-14B, Qwen3.5-4B, Qwen3.5-9B, Ministral-3.3B, Mistral-7B, Llama-3-8B, and Gemma-3-12B---at target sparsities of 40\%, 50\%, 60\%, and 70\%, producing 40 threshold configurations in total.

For each layer and each calibration signal, we maintain a reservoir of at most 200,000 activation samples and obtain the threshold directly from the corresponding empirical quantile.
We additionally save a 4,096-bin activation histogram for each layer to support analysis and reproducibility.
Threshold values are determined separately for each layer, and no additional layer-wise sparsity profile is imposed: within a configuration, all layers share the same target $s_1$ and $s_2$, while their numerical thresholds are determined by their own activation distributions.
All evaluations use the saved calibration artifacts; thresholds are neither re-estimated nor updated during inference.

\textbf{Sequential layer-wise calibration.}
A direct approach would calibrate every layer independently using activations collected from the dense model, but this creates a mismatch with sparse inference: sparsification in earlier layers changes the input distribution observed by later layers.
Prox therefore calibrates layers in network-depth order so that each layer observes the effects of all previously finalized sparse layers.
Algorithm~\ref{alg:sequential_threshold_calibration} summarizes the procedure.

\begin{algorithm}[t]
  \caption{Sequential threshold calibration}
  \label{alg:sequential_threshold_calibration}
  \begin{algorithmic}[1]
    \REQUIRE Calibration set $\mathcal{D}$ and a model with $L$ FFN layers
    \FOR{$\ell=1,\ldots,L$}
    \STATE Run $\mathcal{D}$ through layers $1,\ldots,\ell-1$ using their finalized sparse rules
    \STATE Collect the current-layer inputs $\mathbf{x}_{\ell}$
    \STATE Choose $\tau_{\mathbf{x},\ell}^{s_1}$ and construct $\mathbf{m}_{\mathbf{x},\ell}$
    \STATE Compute $\tilde{\mathbf{s}}_{\ell}$ using $\mathbf{m}_{\mathbf{x},\ell}$ and the quantized proxy weights
    \STATE Choose $\tau_{\mathbf{s},\ell}^{s_2}$ and finalize the sparse rule for layer $\ell$
    \ENDFOR
  \end{algorithmic}
\end{algorithm}

Specifically, when calibrating layer $\ell$, Prox first collects its FFN input $\mathbf{x}_{\ell}$ and determines $\tau_{\mathbf{x},\ell}^{s_1}$.
It then applies the resulting input mask and the model-specific symmetric-INT4 proxy weights to obtain
\[
  \tilde{\mathbf{s}}_{\ell}
  =\widetilde{W}_{\mathrm{up},\ell}
  (\mathbf{m}_{\mathbf{x},\ell}\odot\mathbf{x}_{\ell})
  \odot
  \mathrm{SiLU}\!\left(
    \widetilde{W}_{\mathrm{gate},\ell}
    (\mathbf{m}_{\mathbf{x},\ell}\odot\mathbf{x}_{\ell})
  \right),
\]
from which it determines $\tau_{\mathbf{s},\ell}^{s_2}$.
All preceding FFNs use their finalized Stage~1 and Stage~2 thresholds, whereas the current layer remains dense while its input activations are collected.
Consequently, each layer's calibration distribution incorporates the error propagation and distribution shift induced by earlier sparse layers and more closely matches runtime sparse inference.

\textbf{Runtime sparsity.}
Prox uses fixed magnitude thresholds rather than per-token top-$k$ selection; consequently, the realized mask sparsities can differ from their calibration targets under a new input distribution.
Let $\widehat{s}_1$ and $\widehat{s}_2$ denote the Stage~1 and Stage~2 sparsities measured by accumulating runtime mask counts.
Consistent with the cost model in Sec.~\ref{sec:sparsity_tradeoff}, the measured runtime FFN sparsity is
\[
  \widehat{e}
  =\widehat{s}_2-
  \frac{2\alpha(1-\widehat{s}_1)}{3},
  \qquad \alpha=\frac{1}{3}.
\]
This quantity includes both the sparse exact FFN computation and the additional INT4 proxy cost.

We measure $\widehat{e}$ using 100 examples from each of five task datasets: MMLU, HellaSwag, ARC-Easy, ARC-Challenge, and PIQA.
For every model and target configuration, we aggregate the Stage~1 and Stage~2 mask counts across all five datasets before applying the cost model.
Table~\ref{tab:runtime_effective_sparsity} reports the resulting mean over the ten evaluated models.

\begin{table}[t]
  \centering
  \small
  \setlength{\tabcolsep}{6pt}
  \renewcommand{\arraystretch}{1.08}
  \begin{tabular}{@{}cc@{}}
    \toprule
    \textbf{Target sparsity $e$}
    & \textbf{Mean measured sparsity $\widehat{e}$} \\
    \midrule
    40\% & 38.71\% \\
    50\% & 48.74\% \\
    60\% & 58.44\% \\
    70\% & 68.33\% \\
    \bottomrule
  \end{tabular}
  \caption{Target and runtime FFN sparsities for Prox. Runtime values aggregate five task datasets and are averaged over ten models, with 100 examples evaluated per dataset.}
  \label{tab:runtime_effective_sparsity}
\end{table}

Across all four operating points, the mean measured sparsity is slightly below the target, with gaps of 1.26--1.67 percentage points and no sharp increase at higher target sparsity.
These results indicate that static thresholds calibrated on generic instruction text preserve the intended compute level reasonably well across the evaluated models and task datasets.

\subsection{Perplexity Results}
\label{app:ppl_results}

For this evaluation, we use the same 128 randomly sampled sequences from the WikiText validation set \citep{wikitext} for all models, with a 2,048-token context and a 512-token evaluation window.
We sparsify the full sequence so that the hidden states scored for perplexity are computed under sparsity.
Because perplexity is computed over each model's native token space, Gemma-3-12B's substantially larger vocabulary and different tokenization granularity make its absolute perplexity values not directly comparable with those of the other model families.

\begin{table*}[t]
  \centering
  \footnotesize
  \setlength{\tabcolsep}{0pt}
  \renewcommand{\arraystretch}{1.08}
  \begin{tabular}{@{}L{0.135\textwidth}C{0.095\textwidth}*{10}{C{0.0765\textwidth}}@{}}
    \toprule
    & & \multicolumn{4}{c}{\textbf{Qwen3}}
    & \multicolumn{2}{c}{\textbf{Qwen3.5}}
    & \multicolumn{1}{C{0.0765\textwidth}}{\textbf{\shortstack{Ministral}}}
    & \multicolumn{1}{C{0.0765\textwidth}}{\textbf{Mistral}}
    & \multicolumn{1}{C{0.0765\textwidth}}{\textbf{Llama3}}
    & \multicolumn{1}{C{0.0765\textwidth}}{\textbf{Gemma3}} \\
    \cmidrule(lr){3-6}
    \cmidrule(lr){7-8}
    \cmidrule(lr){9-9}
    \cmidrule(lr){10-10}
    \cmidrule(lr){11-11}
    \cmidrule(lr){12-12}
    \textbf{Variant} & \textbf{Sparsity} & \textbf{1.7B} & \textbf{4B} & \textbf{8B} & \textbf{14B} & \textbf{4B} & \textbf{9B} & \textbf{3.3B} & \textbf{7B} & \textbf{8B} & \textbf{12B} \\
    \midrule
    Dense & 0 & 14.7 & 11.3 & 8.2 & 7.2 & 8.0 & 7.4 & 8.5 & 4.9 & 7.1 & 7.4 \\
    \midrule
    CATS & 40\% & 29.8 & 15.1 & 9.1 & 7.8 & 10.4 & 8.8 & 14.8 & 6.8 & 8.6 & 34.0 \\
    COUNTDOWN & 40\% & 22.0 & 14.6 & 9.8 & 8.2 & 9.5 & 8.3 & 11.4 & 5.3 & 8.9 & \textbf{31.1} \\
    TEAL & 40\% & \textbf{14.3} & 11.7 & \textbf{8.3} & 7.3 & \textbf{8.3} & \textbf{7.5} & \textbf{8.9} & \textbf{5.0} & \textbf{7.3} & 807.0 \\
    Prox & 40\% & 14.5 & \textbf{11.5} & \textbf{8.3} & \textbf{7.2} & \textbf{8.3} & 7.6 & \textbf{8.9} & \textbf{5.0} & 7.5 & 36.0 \\
    \midrule
    CATS & 50\% & 98.3 & 61.6 & 15.1 & 11.6 & 15.8 & 12.2 & 95.3 & 35.5 & 17.4 & 221.8 \\
    COUNTDOWN & 50\% & 47.7 & 20.8 & 12.7 & 10.4 & 12.5 & 10.6 & 16.9 & 6.4 & 12.6 & 254.7 \\
    TEAL & 50\% & \textbf{14.4} & 12.4 & 8.7 & 7.5 & 8.6 & 7.7 & 9.4 & 5.2 & \textbf{7.8} & 898.8 \\
    Prox & 50\% & 14.7 & \textbf{11.8} & \textbf{8.4} & \textbf{7.3} & \textbf{8.5} & \textbf{7.6} & \textbf{9.3} & \textbf{5.1} & \textbf{7.8} & \textbf{185.0} \\
    \midrule
    CATS & 60\% & $1.7{\times}10^{3}$ & $3.6{\times}10^{4}$ & $1.3{\times}10^{3}$ & 521.5 & 121.7 & 56.7 & $9.7{\times}10^{4}$ & $7.5{\times}10^{4}$ & $2.3{\times}10^{5}$ & $1.6{\times}10^{4}$ \\
    COUNTDOWN & 60\% & $1.6{\times}10^{3}$ & 56.5 & 34.1 & 21.9 & 36.6 & 24.4 & 64.6 & 13.5 & 40.2 & \textbf{712.2} \\
    TEAL & 60\% & 17.6 & 14.3 & 9.6 & 8.0 & 9.6 & 8.2 & 11.0 & 5.9 & 8.9 & $1.2{\times}10^{3}$ \\
    Prox & 60\% & \textbf{15.0} & \textbf{12.2} & \textbf{8.6} & \textbf{7.4} & \textbf{8.6} & \textbf{7.8} & \textbf{9.7} & \textbf{5.3} & \textbf{8.2} & 946.4 \\
    \midrule
    TEAL & 70\% & 48.8 & 24.5 & 16.1 & 10.0 & 13.2 & 10.3 & 16.8 & 8.6 & 13.5 & $2.9{\times}10^{3}$ \\
    Prox & 70\% & \textbf{15.7} & \textbf{13.9} & \textbf{9.4} & \textbf{7.8} & \textbf{9.4} & \textbf{8.2} & \textbf{11.6} & \textbf{5.8} & \textbf{9.5} & $\mathbf{1.2{\times}10^{3}}$ \\
    \bottomrule
  \end{tabular}
  \caption{WikiText perplexity across model families and sparsity levels. Lower scores are better.}
  \label{tab:ppl}
\end{table*}

Table~\ref{tab:ppl} reports WikiText perplexity across model families and sparsity levels.
At 40\% sparsity, Prox remains competitive with TEAL.
From 50\% sparsity onward, Prox matches or outperforms TEAL on most model families; at 60\% and 70\%, it achieves lower perplexity on at least nine of the ten evaluated models, with the gap widening markedly at 70\%.
CATS and COUNTDOWN degrade earlier, frequently producing large perplexity increases at 50\% and 60\% sparsity.

Despite its large perplexity increase under sparsification, Gemma-3-12B retains strong downstream accuracy.
Consistent with our observation, \citet{universal_properties_activation_sparsity} report that Gemma-3-12B retains approximately 99\% of its dense average downstream accuracy at 78.77\% activation sparsity.

\subsection{Sensitivity to the Fixed Stage~1 Sparsity}
\label{app:a2_sweep}

Table~\ref{tab:a2_sweep} compares Prox with fixed Stage~1 sparsities from 0.5 to 0.9 on Qwen3-8B to test the sensitivity of A2's $s_1=0.8$ setting.
For each fixed $s_1$ and target effective sparsity $e$, we adjust $s_2$ using the cost model in Sec.~\ref{sec:sparsity_tradeoff} to match the same compute budget and report the aggregate downstream score under the main six-task protocol.

\begin{table}[t]
  \centering
  \footnotesize
  \setlength{\tabcolsep}{4.0pt}
  \renewcommand{\arraystretch}{1.08}
  \begin{tabular*}{\columnwidth}{@{\extracolsep{\fill}}ccccc@{}}
    \toprule
    & \multicolumn{4}{c}{\textbf{Target effective sparsity $e$}} \\
    \cmidrule(lr){2-5}
    \textbf{Setting} & \textbf{40\%} & \textbf{50\%} & \textbf{60\%} & \textbf{70\%} \\
    \midrule
    Prox & 75.7 & 74.6 & 73.3 & 68.6 \\
    \midrule
    $s_1=0.5$ & 73.6 & 73.6 & 72.4 & 65.9 \\
    $s_1=0.6$ & \textbf{75.3} & 74.3 & 72.4 & 65.8 \\
    $s_1=0.7$ & 73.3 & 73.9 & \textbf{72.5} & \textbf{68.6} \\
    $s_1=0.8$ & 74.4 & \textbf{74.3} & 68.9 & 65.7 \\
    $s_1=0.9$ & 69.5 & 68.3 & 67.1 & 58.4 \\
    \bottomrule
  \end{tabular*}
  \caption{Aggregate downstream scores on Qwen3-8B for Prox and fixed Stage~1 sparsities. Higher is better; bold marks the best fixed allocation for each target effective sparsity.}
  \label{tab:a2_sweep}
\end{table}

The best fixed allocation changes with the target: $s_1=0.6$ performs best at 40\% effective sparsity, $s_1=0.8$ at 50\%, and $s_1=0.7$ at 60--70\%.
Prox matches or outperforms every fixed setting at all four targets; even the strongest single fixed strategy, $s_1=0.7$, has a lower average across the four targets (72.1 versus 73.1).
At 70\%, overly aggressive Stage~1 sparsity also reduces the aggregate downstream score from 68.6 at $s_1=0.7$ to 58.4 at $s_1=0.9$.
Thus, no fixed Stage~1 allocation is uniformly best; target-aware allocation avoids per-target downstream sweeps.

\subsection{Sensitivity to the Stage~1 Sparsity Cap}
\label{app:stage1_cap_sensitivity}

Table~\ref{tab:stage1_cap_sensitivity} sweeps the maximum Stage~1 sparsity $s_{1,\max}$ across Qwen3-4B, Qwen3-8B, and Qwen3-14B.
For every setting, we clamp $s_1$ using the corresponding cap and adjust $s_2$ according to the cost model in Sec.~\ref{sec:sparsity_tradeoff} to preserve the target effective sparsity.

The cap affects only the high-sparsity operating points: at 40\% and 50\% effective sparsity, the unconstrained allocation is feasible for every tested cap, while at 60\% only the $s_{1,\max}=50\%$ setting changes the allocation.
At 70\% effective sparsity, all tested caps are active, with $s_2$ adjusted to satisfy the same effective-sparsity budget.

\begin{table}[t]
  \centering
  \footnotesize
  \setlength{\tabcolsep}{3.1pt}
  \renewcommand{\arraystretch}{1.08}
  \begin{tabular*}{\columnwidth}{@{\extracolsep{\fill}}ccc rrr@{}}
    \toprule
    & & & \multicolumn{3}{c}{\textbf{Qwen3}} \\
    \cmidrule(lr){4-6}
    \shortstack{\textbf{Stage~1 cap}\\\textbf{$s_{1,\max}$}} & \shortstack{\textbf{Effective}\\\textbf{sparsity}} & \shortstack{\textbf{Stage~2}\\\textbf{sparsity $s_2$}} & \textbf{4B} & \textbf{8B} & \textbf{14B} \\
    \midrule
    90\% & 70\% & 72.2\% & 46.8 & 57.9 & 72.3 \\
    80\% & 70\% & 74.4\% & 53.7 & 66.6 & 70.7 \\
    70\% & 70\% & 76.7\% & \textbf{58.7} & \textbf{68.6} & \textbf{74.8} \\
    60\% & 70\% & 78.9\% & 58.0 & 65.7 & 72.4 \\
    50\% & 60\% & 71.1\% & 66.3 & 71.0 & 76.4 \\
    50\% & 70\% & 81.1\% & 56.6 & 66.3 & 72.3 \\
    \bottomrule
  \end{tabular*}
  \caption{Sensitivity to the maximum Stage~1 sparsity $s_{1,\max}$ across Qwen3 models. Entries are aggregate downstream scores; each setting adjusts Stage~2 sparsity $s_2$ to match the listed effective sparsity. Higher is better; bold marks the best 70\% setting for each model.}
  \label{tab:stage1_cap_sensitivity}
\end{table}

The $s_{1,\max}=70\%$ cap achieves the highest aggregate downstream score at 70\% effective sparsity for all three model scales.
Increasing the cap to 90\% reduces the scores of Qwen3-4B and Qwen3-8B from 58.7 to 46.8 and from 68.6 to 57.9, respectively.
We therefore use $s_{1,\max}=70\%$ as a shared cap across models without per-model downstream tuning.

\subsection{Detailed Analysis with TEAL}
\label{app:teal_analysis}

\begin{figure}[t]
  \centering
  \includegraphics[width=\columnwidth]{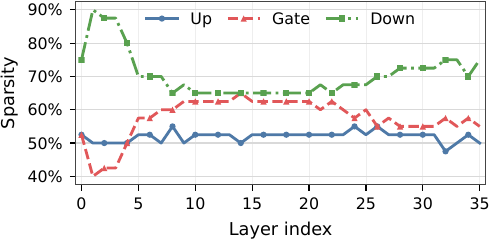}
  \caption{Layer-wise sparsity assigned by TEAL's greedy calibration on Qwen3-8B at 60\% effective sparsity. At every layer, the up-, gate-, and down-projection sparsities average to 60\%.}
  \label{fig:teal_greedy_opt}
\end{figure}

TEAL's block-wise greedy calibration assigns sparsity separately to the up, gate, and down projections while preserving the target effective sparsity.
Starting from dense computation, it repeatedly increases the sparsity of the projection whose candidate update incurs the smallest block-output $\ell_2$ error on calibration activations, until the target block-level budget is reached.
Figure~\ref{fig:teal_greedy_opt} shows the resulting layer-wise allocation for Qwen3-8B at 60\% effective sparsity.

The allocation is both projection- and depth-dependent.
In the first four layers, the down-projection sparsity reaches 75--90\%, compared with 50--52.5\% for the up projection and 40--52.5\% for the gate projection.
Beyond these early layers, the three rates become closer, but the down projection generally remains the most sparse and the up projection the least sparse.
Thus, TEAL's calibrated 60\% operating point does not correspond to applying a uniform sparsity rate to every projection and layer; it instead reveals that the down projection tolerates more approximation, whereas the up projection is generally more sensitive.

This fine-grained search makes TEAL a stronger reference than a uniformly allocated baseline.
In contrast, Prox uses a simple target-aware heuristic motivated by the oracle-$\mathbf{s}$ results: it treats 70\% Stage~2 channel sparsity as a quality-preserving anchor and assigns the remaining budget to Stage~1, using the same allocation across layers.
Despite TEAL's greater allocation flexibility, Prox scores 73.3 versus TEAL's 71.4 on Qwen3-8B at the matched 60\% operating point and outperforms TEAL on all ten evaluated models at both 60\% and 70\% sparsity in Table~\ref{tab:downstream}.
Together with the exact-computation ablation in Table~\ref{tab:ablation}, this result supports the central two-stage design: the proxy is used only to select channels, while the retained values are computed with the original weights, preventing proxy-value errors from propagating directly through the final FFN computation.

The branch-wise allocation provides a further insight into proxy construction.
The lower sparsity generally assigned to the up projection suggests that preserving its computation is more important under TEAL's block-error objective.
This trend is consistent with the comparison between CATS and COUNTDOWN: CATS densely computes the gate-branch signal $\mathbf{h}$ for channel selection, whereas COUNTDOWN uses the densely computed up-branch signal $\mathbf{u}$; across the 30 matched model--sparsity settings at 40--60\% in Table~\ref{tab:downstream}, COUNTDOWN always achieves a higher aggregate score than CATS.
Although projection sensitivity and selection-signal quality are not identical quantities, these results jointly suggest that $\mathbf{u}$ provides a more informative approximation to the joint intermediate signal $\mathbf{s}$ than $\mathbf{h}$ alone.

Prox currently uses the same sparsity allocation across all layers.
A promising extension is to follow TEAL's layer-wise allocation and use different Stage~1 sparsities when constructing the $\mathbf{u}$- and $\mathbf{h}$-branch proxies, which may improve the accuracy of proxy-based channel selection.


\end{document}